%% file: main.tex
\documentclass[letterpaper]{article} 
\usepackage[final,nonatbib]{neurips_2023}
\usepackage[numbers, sort, compress]{natbib}
\usepackage{times}  
\usepackage{helvet}  
\usepackage{courier}  
\usepackage[hyphens]{url}  
\usepackage{graphicx} 
\usepackage{caption} 
\usepackage{algorithm}
\usepackage{algorithmic}
\usepackage{adjustbox}

\usepackage{newfloat}
\usepackage{listings}
\DeclareCaptionStyle{ruled}{labelfont=normalfont,labelsep=colon,strut=off} 
\floatstyle{ruled}
\newfloat{listing}{tb}{lst}{}
\floatname{listing}{Listing}
\title{\workingtitle}

\author{%
  Kweku Kwegyir-Aggrey \\ 
  Brown University\\
  \texttt{kweku@brown.edu} \\
    \And
  A. Feder Cooper \\ 
  Cornell University\\
    \texttt{afc78@cornell.edu} \\
    \And
  Jessica Dai \\ 
  UC Berkeley\\
\texttt{jessicadai@berkeley.edu} \\
 \And
  John P. Dickerson\\ 
  Arthur AI\\
\texttt{john@arthur.ai} \\
\And
  Keegan Hines\\ 
  Arthur AI\\
\texttt{keegan@arthur.ai} \\
\And
  Suresh Venkatasubramanian \\
  Brown University \\
  \texttt{suresh\_venkatasubramanian@brown.edu}\\
}

\usepackage{bibentry}

\usepackage{bbold}
\usepackage{bm}
\usepackage{amsmath,amsthm}
\usepackage{algorithm}
\usepackage{multirow}
\usepackage{amsmath}
\usepackage{subcaption} 
\usepackage{dsfont} 
\usepackage{mathtools}
\usepackage{amssymb}
\usepackage{nameref}
\usepackage{graphicx}
\usepackage[abs]{overpic}

\usepackage{algorithm}

\usepackage{hhline}
\usepackage[utf8]{inputenc} 
\usepackage[T1]{fontenc}    
\usepackage{url}            
\usepackage{booktabs}       
\usepackage{amsfonts}       
\usepackage{nicefrac}       
\usepackage{microtype}      
\usepackage{xcolor}         
\usepackage{enumitem}       

\usepackage{multirow}
\usepackage{adjustbox,lipsum}

\newtheorem{assumption}{Assumption}[section]
\newtheorem{definition}{Definition}[section]
\newtheorem{theorem}{Theorem}[section]
\newtheorem{proposition}{Proposition}[section]
\newtheorem{remark}{Remark}[section]
\newtheorem{corollary}{Corollary}[section]

\newtheorem{lemma}{Lemma}[section]

\usepackage{theory}
\usepackage{macros}

\begin{document}

\maketitle

\input{sections/00_abstract}
\input{sections/10_introduction}

\input{sections/11_background}
\input{sections/17_formulation_alt} 
\input{sections/41_geometric}
\input{sections/45_algorithm}

\input{sections/50_experiments}

\input{sections/25_related_work.tex}
\input{sections/60_conclusion}

\bibliography{references}
\bibliographystyle{plainnat}
\input{sections/appendix/appendix.tex}

\end{document}

%% file: sections/00_abstract.tex
\begin{abstract}
We study the problem of post-processing a supervised machine-learned regressor to maximize fair binary classification at all decision thresholds. By decreasing the statistical distance between each group's score distributions, we show that we can increase fair performance across all thresholds at once, and that we can do so without a large decrease in accuracy. To this end, we introduce a formal measure of \emph{Distributional Parity}, which captures the degree of similarity in the distributions of classifications for different protected groups. Our main result is to put forward a novel post-processing algorithm based on optimal transport, which provably maximizes Distributional Parity, thereby attaining common notions of group fairness like Equalized Odds or Equal Opportunity at all thresholds. We demonstrate on two fairness benchmarks that our technique works well empirically, while also outperforming and generalizing similar techniques from related work. \looseness=-1 
\end{abstract}

%% file: sections/10_introduction.tex
\section{Introduction}\label{sec:introduction}
A common approach to fair machine learning is to train a classifier with a chosen decision threshold in order to attain a certain degree of accuracy, and then to post-process the classifier to correct for unfairness according to a chosen fairness definition \cite{calders2009demographicparity, hardt2016equality, pleiss2017fairness}. Despite the popularity of this approach, it suffers from two major limitations. First, it is well-known that the specific choice of decision threshold can influence both fairness and accuracy in practice \cite{barocas2019book} producing an undesirable trade-off between the two objectives. Second, when deploying a classifier in the real world, practitioners typically need to tinker with the threshold as they evaluate whether a model meets their domain-specific needs \cite{kallus2019fairness, chouldechova2016fair}.
One strategy to address these limitations is to develop a procedure that produces regressors that guarantee a selected fairness notion at \emph{all} possible thresholds, while simultaneously preserving accuracy. If a regressor is fair at all thresholds, then a practitioner can freely perform application-specific threshold tuning without ever needing to retrain.  

Some prior work has investigated this strategy, by using optimal-transport methods to achieve a single, often trivially satisfied, fairness notion -- Demographic Parity -- at all thresholds~ \cite{jiang2020wasserstein, legouic2020projection, chzhen2020fairwasserstein}. 
However, \cite{hardt2016equality} and related impossibility results~\cite{kleinberg2018impossibility, chouldechova2016fair} demonstrate that attaining fairness only in the sense of Demographic Parity does \emph{not} capture the nuances in unfairness arising from examining true positive rates, false positive rates, and combinations thereof~\cite{barocas2019book}. We therefore ask: \textit{Can we train a regressor once and obtain fair binary classifiers at all thresholds for more flexible group fairness notions?}\looseness=-1

\paragraph{Our Work.} Our key insight is to observe that parity at the distribution level of a regressor's output for each sensitive group, prior to the application of a threshold, can be harnessed to attain fairness at all thresholds simultaneously. This insight yields the following contributions:
\textbf{(1)} We introduce a metric called \emph{Distributional Parity} (Definition~\ref{def:distributionalparity}) based on the Wasserstein-1 Distance, which enables reasoning about fairness across all thresholds for a wide class of metrics. \textbf{(2)} We employ a technique called Geometric Repair~\cite{feldman2015certifying}, which leverages an important connection to Wasserstein-2 barycenters to post-process a regressor under a  Distributional Parity constraint, attaining all-threshold fairness. \textbf{(3)} We prove that Distributional Parity is convex on the set of models produced by Geometric Repair, thereby enabling efficient computation of our proposed post-processing. Additionally,  we show that the models produced by geometric repair are Pareto optimal in the multi-objective optimization of accuracy (via an $\ell_1$-type risk) and Distributional Parity. \textbf{(4)} Lastly, we synthesize these insights into a novel post-processing algorithm for a broad class of fairness metrics; our algorithm subsumes earlier work on all-threshold Demographic Parity, and we demonstrate its efficacy in experiments on common benchmarks.

%% file: sections/11_background.tex
\vspace{-.3cm}
\section{Background}
\label{sec:background}
\vspace{-.2cm}
Let $\X \subseteq \reals{d}$ be a feature space and $G = \{a,b\}$ be a set of
binary protected attributes for which $\pa$ is the majority group and $\protp$
is the minority group. We denote the label space as $Y = \binaryset$, where $0$
denotes the negative class and $1$ the positive class. We assume elements
in $X$, $G$, and $Y$ are drawn from some underlying distribution, with
corresponding random variables $\Xrv$, $\Protrv$, and $\Yrv$. The proportion of
each group is represented $\rho_\prot = \Pr[\Protrv = \prot]$. 
Let $\mc F$ be a set of measureable \emph{group-aware regressors} (from which binary classifiers are derived). Each regressor $\f \in \mc F$ has signature $f:\X \times \Prot \rightarrow \ospace$ and outputs a \emph{score} $\score \in \ospace$ where $s = \Pr[\Yrv=1|\Xrv =x, \Protrv=\prot]$. For a fixed regressor $f \in \mc F$ and a decision threshold $\tau \in \binaryrange$, we can derive a binary classifier from $f$ by computing $\mathbb{1}\{f \geq \tau\}$ for any $\tau \in \binaryrange$. For a group $g \in \Prot$, the \emph{group-conditional score distribution} is the distribution of scores produced by a regressor on that group.  We denote this distribution $f(\Xrv, \Protrv)|\Protrv=\prot$. 

For $p\geq 1$, we define 
$\Ppm{p}(\ospace)$ to be the space of probability measures on $\ospace$ with finite
$p$-order-moments.  We use $\mu_\prot \in
\Ppm{p}(\ospace)$ to denote the probability measure associated with each group's
score distribution. 
We also make the following (standard) assumption on these measures. 
\begin{assumption}
\label{assumption:wass}
Any measure $\mu \in \Ppm{p}(\ospace)$ with finite $p$-order moments is non-atomic and absolutely continuous with respect to the Lebesgue measure.
\end{assumption}

\noindent This assumption provides two guarantees. First, it ensures the cumulative distribution function (CDF) of $\mu_{\prot}$ denoted $F_{\mu_\prot}(\tau) = \mu_{\prot}([0,\tau))$ has a well defined inverse $F^{-1}_{\mu_\prot}$. Second, it ensures that certain optimal transport operations, upon which our contributions crucially rely, are well-defined.

\vspace{-.2cm}
\subsection{Wasserstein Distance and Wasserstein 
Barycenters}
\label{sec:setting}
\vspace{-.2cm}

Before introducing our solution, we present some necessary background on Wasserstein distance and Wasserstein barycenters.\looseness=-1 
.   

\paragraph*{Wasserstein Distance.} Informally, the Wasserstein distance captures the difference between probability measures by measuring the \textit{cost} of transforming one probability measure into the other.   In the special case when distributions are univariate, the Wasserstein distance has a nice closed form.\looseness=-1 
\begin{definition}[Wasserstein Distance]
 For two measures $\mu_1, \mu_2 \in \Ppm{p}(\ospace)$ 
\vspace{-.1cm}
\begin{align}
\label{eq: wasserstein_def}
    \Wass_{p}^{p}(\mu_1,\mu_2) &= \int_{\ospace}|F^{-1}_{\mu_1}(q) - F^{-1}_{\mu_2}(q)|^p dq.
\end{align} 
\vspace{-.5cm}
\end{definition}
We can also define the Wasserstein distance using transport plans; this is commonly referred to as Monge's Formulation.  A transport plan is a  function $T \in \mc T$ where every function in $\mc T$ satisfies standard pushforward constraints, i.e, $T_{\#}\mu_1 = \mu_2$ such that $\mu_{2}(B) = \mu_1(T^{-1}(B))$ for all measurable $B \subseteq \ospace$. 
\begin{definition}[Wasserstein Distance \text{[Monge]}]
\begin{align}
\vspace*{-.65cm}
\label{eq: wass_tp_def}
     \Wass_{p}^{p}(\mu_1,\mu_2) &= \inf_{T \in \mc T} \int_{\ospace}|q - T(q)|^p d\mu_1(q). 
\end{align} 
\vspace*{-.6cm}
\end{definition}

In our specific case where Assumption \ref{assumption:wass} is satisfied, we know that these transport plans which solve Monge's formulation exist and that we can define them in closed form. 
\begin{remark}
\label{rem:transport_plan}
 The transport plan from $\mu_1 \rightarrow \mu_2$ which minimizes Eq. (\ref{eq: wass_tp_def}) is defined $T_1^2(x) = F^{-1}_{\mu_2}(F_{\mu_1}(x))$ for all $p \geq 1$ \cite[Remark 2.6]{santambrogio2015optimal}  
\vspace*{-.25cm}
\end{remark}
\label{eq:transport_plan}



\paragraph*{Wasserstein Barycenter.} The Wasserstein barycenter 
is a weighted composition of two distributions, much like a weighted average or midpoint in the Euclidean sense; it provides a principled way to compose two measures. 
\begin{definition}[Wasserstein Barycenter]
\label{def:barycenter}
For two measures $\mu_1,\mu_2 \in \Ppm{p}(\ospace)$ their $\alpha$-weighted \textit{Wasserstein barycenter}
is denoted $\mu_\alpha$ and is computed
\begin{align}
\label{eq:barycenter}
\mu_\alpha \leftarrow \argmin_{\nu \in \Ppm{p}(\ospace)}  (1-\alpha)\Wass_p^p(\mu_1, \nu) + \alpha\Wass_p^p(\mu_2, \nu), 
\end{align}
and in the special case when $\alpha=\rho_\protp$ we write $\mu^*$.  
\end{definition}
\noindent To complete the weighted-average analogy, $\alpha$ behaves like a tunable knob: As $\alpha \rightarrow 0$ then $\mu_\alpha$ will appear more like $\mu_1$, and as $\alpha \rightarrow 1$ the more $\mu_\alpha$ will appear like $\mu_2$.  
As a consequence of this definition and Remark \ref{rem:transport_plan}, we can express the transport plan to a barycenter in closed form, as well:\looseness=-1 
\begin{corollary}
\label{cor:barycenter_transport_plan}
Let $\mustar$ be the $\rho_\protp$-weighted barycenter of $\mu_\pa,\mu_\protp$ then the transport plan from $\mu_\pa \rightarrow \mustar$ (wlog) is computed 
$T_\pa^*(\omega) =  (\rho_\pa F^{-1}_{\mu_\pa} + \rho_\protp F^{-1}_{\mu_\protp}) \circ F_{\mu_\pa}(\omega)$. 
\vspace*{-.2cm}
\end{corollary}

\paragraph*{A Note on Our Use of $\Wass_1$ and $\Wass_2$.}  In this work,  we make use of both $\Wass_1$ and $\Wass_2$. Our use of $\Wass_1$ is restricted to Distributional Parity computations (see Section \ref{sec: distributional_parity}). This choice is motivated by the fact when $\gamma=\UF_\PR$, the Wasserstein-1 distance recovers $\UF_\PR$. We use $\Wass_2$ to compute Wasserstein barycenters. Given that $\Wass_2$ is known to be strictly convex, and provided that some $\mu_\prot$ is non-atomic, for $p=2$ the barycenter that minimizes  Eq. \ref{def:barycenter} is unique \cite[Proposition 3.5]{agueh2011barycenters}.  

%% file: sections/17_formulation_alt.tex
\vspace*{-.2cm}
\section{A Distributional View of Fairness}
\vspace*{-.2cm}

\label{sec: distributional_parity}

Our goal is to post-process a regressor such that all binary classifiers derived from thresholding this regressor are group fair, i.e., attain fairness in the regressor at every threshold.  
To attain fairness at every threshold, we look to create parity in outcomes at the level of the regressor -- before thresholds are applied -- rather than at the level of the derived predictor.  The intuition is simple: if a regressor outputs similar scores for two groups, then no matter what threshold is selected, the output from the derived predictor will be fair. 
Specifically, we show that fairness can be attained at all thresholds by enforcing parity in the \emph{distribution} of scores output by a regressor on some groups.

At the core of our new distributional definition of fairness are familiar metrics, namely: Positive Rate (PR), True Positive Rate (TPR), and False Positive Rate (FPR). From these metrics, we can obtain popular fairness definitions, such as Demographic Parity (PR Parity)~\cite{calders2009demographicparity}, Equal Opportunity (TPR Parity), and Equalized Odds (TPR and FPR Parity)~\cite{hardt2016equality}.  

Let the set of these metrics be $\Gamma = \{\PR, \TPR, \FPR\}$ and any arbitrary metric be $\gamma \in \Gamma$. We write $\gamma_\prot(\tau;f)$ to denote the rate $\gamma$ on group $\prot$ at threshold $\tau$ for a score distribution produced by $f$.  When obvious from context, we omit $\f$ from this $\gamma$ notation, writing only $\gamma_\prot(\tau)$.  Additionally, as we show via Corollary \ref{cor:additive}, we can combine these metrics additively, e.g., producing Equalized Odds which combines $\TPR$ and $\FPR$.\looseness=-1

At a single threshold, (un)fairness is commonly measured by taking the difference in some metric across groups --- e.g., for the case of Demographic Parity where $\gamma = \PR$, we can measure fairness by computing $|\PR_{\pa}(\tau) - \PR_{\protp}(\tau)|$.  A natural way to leverage these single-threshold measurements into an all-threshold measurement is to take their average across every possible $\tau$.
We formalize this idea in the following definition of \emph{Distributional Parity}. 
\begin{definition}[Distributional Parity]
\label{def:distributionalparity}
Let $U(\ospace)$ be the uniform distribution on $\ospace$. For a fairness metric $\gamma \in \Gamma$, a regressor $f$ satisfies \emph{Distributional Parity} denoted $\mc U_\gamma(f) \triangleq  
\Expect_{\tau \sim U(\ospace)}| \gamma_{\pa}(\tau) - \gamma_{\protp}(\tau)|$, when $\mathcal{U}_\gamma(f)=0$.
\end{definition}

A useful property of this definition is that when $\gamma = \PR$, this definition is closely related to the Wasserstein Distance, a distance which is frequently used to measure distance between probability distributions.  

\begin{proposition}
\label{cor:wass_sdp}
\wassspdcorollary
\end{proposition}

It is from this property that Distributional Parity is named.  At its core, Distributional Parity is a way to quantify differences between outcome distributions -- specifically the groupwise score distributions of $f$. This relationship between Distributional Parity -- an all-threshold fairness metric --  and the Wasserstein distance --  a measure of statistical distance --  anchors our proposed shift in focus from thresholds to distributions. 

Next, we introduce our proposed post-processing objective for computing fair regressors under a Distributional Parity constraint.  We will also begin to outline how we efficiently compute this post-processing, and how our solution elegantly addresses fairness-accuracy trade-off concerns.\looseness=-1

\subsection{Distributionally Fair Post-Processing} 
\label{sec:formulation}

Our goal is to post-process a learned regressor $f$, such that it becomes (distributionally) fair while remaining accurate. 
The risk of some other regressor $\fhat$ (with respect to $f$) is  computed  
\begin{align*}
\centering
    \mc R(\fhat) = \|\fhat - f\|_{1} = \Expect|\fhat(\Xrv, \Protrv) - f(\Xrv, \Protrv)|
\end{align*}
Using this definition of risk, a simple fair post-processing objective can be written as follows,
\begin{align}
\label{eq: fair_classification}
    \arg\inf_{\fhat \in \mc F} \mc R(\fhat) \quad \suchthat \quad \UF_\gamma(\fhat) \leq c, \hspace{.5cm} \text{where $c$ is some small constant.}
\end{align}
\vspace{-.5cm}

\paragraph{The special case of \PR.} In the special case where $\gamma = \PR$ and $c=0$, the solution to Eq. \ref{eq: fair_classification} can be computed using a solution based on optimal transport \cite{jiang2020wasserstein}.  In this solution, a learned regressor $f$ is transformed into a new regressor we call $\fsdp$ which provably minimizes risk (with respect to $f$) while attaining Distributional Parity for $\gamma = \PR$, i.e, demographic parity at every threshold.  This all threshold guarantee is attained with minimal impact to risk, as it was shown in \cite{legouic2020projection, chzhen2020fairwasserstein} that $\fsdp$ is the regressor which increases risk the \emph{least} amongst all regressors which satisfy all threshold demographic parity constraints.

\begin{align}
    \fsdp \leftarrow \arg\min \mc R(\cdot) \quad \suchthat \quad \UF_\PR(\fsdp) = 0.
\end{align}

This solution is strict -- it enforces exact demographic parity, which may not always be desired \cite{chzhen2022minimax}. We can address this concern by considering a relaxation of Eq. \ref{eq: fair_classification} which uses a parameter $\lambda$ to balance the trade-off between fairness and accuracy. Specifically, for every $\lambda \in \binaryrange$ there is some $f_\lambda \in \mc F$ which attains $\lambda$-increase in the fairness, in exchange for a $\lambda$-reduction in risk.  We prove the existence of $f_\lambda$ which satisfies this property in the following lemma.  
\begin{lemma}
\label{lemma:opt_geometric_repair}
\geometricrepairfairoptimal
\end{lemma}

The functions $f_\lambda$ are Pareto-optimal: indeed, we show in Theorem  \ref{prop:pareto_prop} that all $\{f_\lambda\}_{\lambda \in \binaryrange}$ are Pareto optimal in the multi-objective minimization of $\mc R$ and $\UF_\PR$.  This means we can view these regressors as being optimally accuracy preserving, while also being fair. 
As a result, the above optimization (with $\gamma = \PR$) can be rewritten simply as
\begin{align}
\label{eq: opt_geometric_repair}
    \arg\min_{\lambda \in \binaryrange} \UF_\gamma(f_\lambda), 
\end{align}
replacing the risk minimization objective with an objective that enforces Distributional Parity, given the aforementioned accuracy-preserving properties of $f_\lambda$.

\paragraph{Extending to other fairness metrics.}

The above approach to achieving all-threshold fairness has two steps: a characterization of a solution that achieves ideal fairness, and then the construction of a space of functions $f_\lambda$ that allow for an optimal fairness-accuracy trade-off.  Unfortunately, this exact result for $\PR$ relies heavily on optimal transport in a way that does not naturally generalize to other fairness measures $\gamma \ne \PR$ without modifying the approach.\looseness=-1

To address this, we provide two key insights. Firstly, we show that $f_\lambda$ can be expressed easily, in closed form. Secondly, we prove that the optimization in equation \eqref{eq: opt_geometric_repair} describes a convex function of $\lambda$ \emph{independent of the choice of $\gamma$}.  This indicates that we can optimize Distributional Parity  \emph{for any metric} by selecting an appropriate $f_\lambda$.  Said differently, for any choice of $\gamma$, we can find the optimal value of $\lambda$ to minimize  $\UF_\gamma(f_\lambda)$. Additionally, we provide empirical evidence that this choice yields an almost perfect minimization of Distributional Parity, thus achieving an all-threshold fairness result as desired. 

%% file: sections/41_geometric.tex
\vspace{-.2cm}
\section{Maximizing Distributional Parity with Geometric Repair}
\vspace{-.2cm}
We now introduce the method and main theorem that we use to perform distributionally fair post-processing. The method, defined in Section \ref{sec:introducing_geometric_repair} is called Geometric Repair \cite{feldman2015certifying, chzhen2022minimax}, and is how we efficiently compute solutions to the objective stated in Equation \ref{eq: opt_geometric_repair}. 
The main theoretical result which enables our approach is stated in Theorem \ref{thm:convex}.
Subsequently, we show in subsection \ref{sec:move_mountains} that we can make use of an elegant transformation to an optimal transport problem in order to achieve approximate distributional parity for all $\gamma$.
 
\vspace{-.2cm}
\subsection{Defining Geometric Repair}
\label{sec:introducing_geometric_repair}
Geometric repair is a technique for constructing a regressor that interpolates between the output of some learned regressor $f$ (assumed to be accurate), and the output of a specific fair function $\fsdp$.
Note, $\fsdp$ must be specifically chosen in order to prove our results, but for ease of exposition, we defer the formal definition of $\fsdp$ to the following subsection. 
\begin{definition}[Geometric Repair]
\label{def:geometricrepair}
We call $\lambda \in \binaryrange$ the repair parameter and define a geometrically repaired regressor $f_{\lambda}$ as $f_{\lambda}(x,\prot) \triangleq (1-\lambda)f(x,\prot) + \lambda \fsdp(x,\prot)$.  
\end{definition}
Geometric repair enumerates a well-structured set of regressors which achieve  $\lambda$-relaxations of $R$ and $\UF_\PR$ as described in Section \ref{sec: distributional_parity}. 
\begin{proposition}
\label{prop:lambda_relxation}
For any $\lambda \in \binaryrange$, a repaired regressor $f_\lambda$ satisfies $R(f_\lambda) = \lambda R(\fsdp)$ and $\UF_\PR(f_{\lambda})  = (1-\lambda)\UF_\PR(f)$.
\end{proposition}

 This is the set of regressors used to maximize distributional parity. The key to computing such a maximization lies in the following theorem, which shows that distributional parity is convex, on the set of repaired regressors. This convexity guarantee certifies our ability to locate the $f_\lambda$, amongst the set of repaired regressors, which \textit{best} minimizes Distributional Parity for \emph{any} $\gamma$.  



\begin{theorem}
\label{thm:convex}
\convexityresult
\end{theorem}

\noindent 
The proof of this theorem crucially depends on the connection between $\fsdp$ and Wasserstein barycenters. In the next section, we leverage this connection to analytically compute the distributions of $f_\lambda$, which is a crucial piece needed in proving the convexity of $\UF_{\gamma}(f_\lambda)$.

\subsection{How $\fsdp$ Enables Geometric Repair} 
\label{sec:move_mountains}

Here, we formalize the earlier definition of $\fsdp$ from Section \ref{sec:move_mountains} and its connection between Wasserstein barycenters in the context of geometric repair.
\begin{definition}[Fully Repaired Regressor]
The regressor $\fsdp$ which satisfies distributional parity for $\gamma=\PR$ while minimizing risk (with respect to $f$) is the computed 
\begin{align}
\label{eq: SDP_processing}
    \fsdp \leftarrow \arg\min_{f \in \mc F} \mc R(\cdot) \quad \suchthat \quad \UF_\PR(f) = 0.
\end{align}
We call this regressor \emph{fully repaired} in that $\f_{\lambda=1}$ is equivalent to $\fsdp$.
\end{definition}

The aforementioned property which relates $\fsdp$ to $\Wass_2$ barycenters is the fairness constraint in Eq. (\ref{eq: SDP_processing}). To make this clear, recall 
Proposition \ref{cor:wass_sdp} which states that removing the $\Wass_2$ distance between distributions is sufficient to satisfy distributional parity for $\gamma=\PR$. The tool we will use to remove this distance is  Wasserstein barycenters.  Prior work \cite{le2017existence, chzhen2020fairwasserstein} show that mapping $\mu_\pa, \mu_\protp$ onto their $\rho_\protp$-weighted barycenter distribution, which we denote $\mustar$, removes the Wasserstein distance between $\mu_\pa,\mu_\protp$ under this mapping, thereby satisfying $\UF_\PR(\fsdp)=0$ and establishing that $\fsdp$ is distributed like $\mustar$. 

\begin{figure}[t!]
    \centering
    \vspace{-2cm}
    \begin{overpic}[abs,unit=1mm,scale=.25]{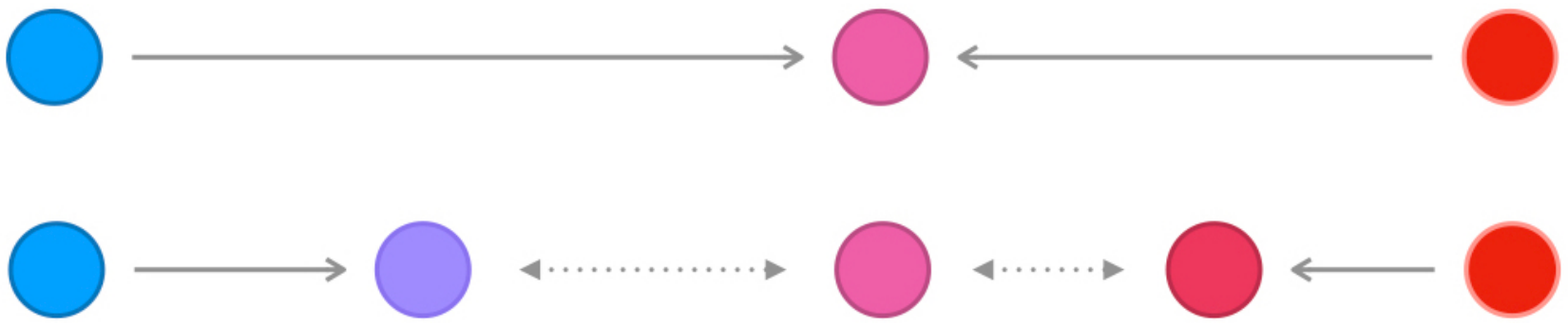}
    \put(1,26.5){$\mu_\pa$}
    \put(66,26.5){$\mu_\protp$}
    \put(38,26.5){$\mustar$}
    \put(16,26.5){$\mu_{\pa,\lambda}$}
    \put(19,33){$\rho_\pa$}
    \put(52,33){$\rho_\protp$}

    \put(52,26.5){$\mu_{\protp,\lambda}$}
    \end{overpic}
    \vspace{-2cm}
    \caption{Let $\mu_\pa,\mu_\prot$ be groupwise score distributions. We illustrate the repaired score distributions $\mu_{\prot,\lambda}$ under geometric repair, where $\mustar$ is the $\rho_\protp$-weighted barycenter. }
    \label{fig:repairtimeline}
    \vspace{-.4cm}
\end{figure}

We can use this fact to rewrite the score distributions of each group under geometric repair. The following proposition formalizes this claim by showing that the groupwise score distributions output by any $f_\lambda$ can be computed as barycenters of $\mu_\prot$ and $\mustar$. \looseness=-1
\begin{proposition}
\label{prop:repaired_barycenters}
\barycenterprop
\end{proposition}
  This proposition shows us that the interpolation between $f$ and $\fsdp$ as parametrized by $\lambda$ in geometric repair is replicated at the distributional level, i.e., $\lambda$ also controls the interpolation from $\mu_{\prot, \lambda} \rightarrow \mustar$; more importantly, the intermediate distributions of this interpolation have a special structure -- they are also barycenters. Note that under Assumption \ref{assumption:wass} and \cite[Proposition 3.5]{agueh2011barycenters}, these $\mu_{\prot,\lambda}$ are unique and guaranteed to exist. For clarity, we visualize this interpolation (over distributions) in Figure \ref{fig:repairtimeline}.\looseness=-1

\noindent Our proof of Theorem \ref{thm:convex} computes distributional parity as a function of the score distributions of $f_\lambda$.  With these established via Proposition \ref{prop:repaired_barycenters}, we can write a closed form expression for $\UF_\gamma(\f_\lambda)$. Using this expression, the proof proceeds computationally, showing that the second derivative of $\UF_\gamma(f_\lambda)$ is non-negative to conclude convexity.

\subsection{The Optimality of Geometric Repair in Balancing Fairness and Accuracy}
Now, we will show that $f_\lambda$ is optimal in the fairness-accuracy trade-off with respect to $\gamma = \PR$.  

\begin{definition}[Pareto Optimality]
For $f, f' \in \mc F$ we say $f$ Pareto dominates $f'$, denoted $f' \prec f$, if one of the following hold:
\begin{align}
\mc R(f) \leq \mc R(f') \quad \UF_\PR(f) < \UF_\PR(f') \\
\mc R(f) < \mc R(f') \quad \UF_\PR(f) \leq \UF_\PR(f')
\end{align} 
\end{definition}
A regressor $f$ is Pareto optimal if there is no other regressor $f'$ that has improved risk without also having strictly more unfairness, or vice-versa.

The proof of Pareto optimality of $f_\lambda$ follows from Proposition \ref{prop:lambda_relxation}. The main idea of this result is the following: $\fsdp$ is the lowest risk classifier where $\UF_\PR(\cdot) =0$ meaning that it is Pareto optimal by construction. Since $f_\lambda$ is a $\lambda$-relaxation of $\fsdp$ with regards to both risk and unfairness, $f_\lambda$ preserves the Pareto optimality of $\fsdp$. 
\begin{theorem}
\label{prop:pareto_prop}
\paretooptimal.
\end{theorem}


%% file: sections/45_algorithm.tex
\vspace{-.2cm}
\section{Post-Processing Algorithms to Maximize Distributional Parity}
\label{section:algorithms}
\vspace{-.2cm}
Now that we have supported \textit{why} we can use geometric repair to maximize distributional parity, we provide some practical algorithms showing \textit{how} to do so. First, we show how to estimate $\f_\lambda$ from samples.\looseness=-1 

\paragraph{Plug-in Estimator for $\f_\lambda$.} Indeed, computation of $\fsdp$, and therefore $\mustar$, requires exact knowledge of $\mu_\pa, \mu_\protp$. In practice, we only have sample access to both score distributions, and so we must approximate these distributions, and consequently their barycenter and $f_\lambda$. We show a plug-in estimator w/ the following convergence guarantee (Theorem \ref{thm:empirical_dist}) to approximate $\f_\lambda$ in Algorithm \ref{alg:make_f_lambda}.  Our approach to approximating $f_\lambda$ only requires an input regressor $f$ and access to some unlabeled dataset $D = (x_1,g_1) \dots (x_n,x_g)$.  Let $n_g$ denote the number of samples from a group $g$. 
\begin{theorem}
\label{thm:empirical_dist}
\empiricaldisttheorem 
\end{theorem}

\paragraph*{Post-Processing to Maximize Distributional Parity. } To actually compute the optimal $\lambdastar$ for some metric, we propose the post-processing routine described in  Algorithm~\ref{alg:repair}. The algorithm consists of two main steps: approximating $\hat{f}_\lambda$ in Step 1 and finding the optimal $\lambdastar$ in Step 2. Note that our objective $\hat{\UF}_{\gamma}(f_\lambda)$ is parametrized by the scalar $\lambda$, and so we find its minima using a univariate solver; we found success using Brent's Method \cite{brent2013algorithms}. By the convexity of $\UF_\gamma(\cdot)$ as proven in Theorem \ref{thm:convex}, we are guaranteed that the $\f_{\lambdastar}$ is optimal on the set of repaired regressors.

\begin{algorithm}[t!]
\caption{An Estimator for $\f_\lambda$}\label{alg:make_f_lambda} 
{\bfseries Input:} A regressor $f$, and an unlabeled dataset $D = (x_1,\prot_1)\dots (x_n,\prot_n)$
   \begin{enumerate}
    \item Let $n_\prot = \frac{1}{n}\sum_{k=1}^{n} \mathbb{1}_{g_k = \prot}$.  Use $f$ to approximate the group-conditional distributions
    \begin{align*}
        \hat{\mu_\prot} = \frac{1}{n_\prot}\sum_{i=1}\delta_{f(x_i, \prot_i)}\mathbb{1}_{\prot_i = \prot} 
    \end{align*}
    \item Let $\hat{\rho}_\prot = \frac{n_\prot}{n}$ and compute the empirical optimal transport plans (see Remark \ref{rem:transport_plan})
    \begin{align*}
        \hat{T}_\prot^*(\omega) =  (\hat{\rho}_\pa F^{-1}_{\hat{\mu}_\pa} + \hat{\rho}_\protp F^{-1}_{\hat{\mu}_\protp}) \circ F_{\hat{\mu}_\pa}(\omega) 
    \end{align*}
    \item For any $\lambda \in \binaryrange$, compute $\hat{f}_\lambda$ where $\hat{f}_\lambda(x,g) = (1-\lambda)f(x,\prot) + \lambda \hat{T}_\prot^*(f(x,\prot)) $
   
\end{enumerate}
\end{algorithm}

\begin{algorithm}[tb]

\caption{Post-Processing for Distributional Parity }\label{alg:repair} 
    {\bfseries Input:} A metric $\gamma \in \Gamma$, learned regressor $f$, and labeled dataset $E = (x_1, \prot_1, y_1) \dots ...(x_k, \prot_k, y_k)$ 
   \begin{enumerate}
    \item Using Algorithm (\ref{alg:make_f_lambda}) to approximate $f_\lambda$ by computing $\hat{T}^*_\prot$ such that for all $\lambda \in \binaryrange$ geometric repair is well defined, i.e., $\hat{f}_\lambda(x,g) = (1-\lambda)f(x,g) + \lambda\hat{T}^*_\prot(f(x,g))$      
    \item Use Brent's algorithm to find the optimal $\lambda$ which minimizes $ \lambda_* \leftarrow Brent_{\lambda \in \binaryrange} \hat{\UF}_{\gamma}(\fhat_\lambda) $ 
    where $\hat{\UF}(f_\lambda)$ is approximated for $m$ randomly sampled $(\tau_1....\tau_m) \sim U(\binaryrange)$ via 
    \begin{align*}
        \hat{\UF}(\fhat_\lambda)= \frac{1}{m}\sum_{\ell = 1}^{m} | \gamma_{\pa}(\tau_\ell; \fhat_\lambda) - \gamma_{\protp}(\tau_\ell; \fhat_\lambda) |.
    \end{align*}
    \item \textbf{Output:} $f_{\lambda_*}(x,g)$ such that $\hat{\UF}_{\gamma}(f_{\lambda_*})$ is minimized (distributional parity is maximized)  
\end{enumerate}
\end{algorithm}

\begin{corollary}
\label{cor:additive}
Since convex functions are closed under addition, Theorem \ref{thm:convex} also applies to additive combinations of metrics, meaning that the objective in Step (2) of Alg \ref{alg:repair} can be replaced by $\UF_{\gamma_1}(f_\lambda) +\UF_{\gamma_2}(f_\lambda) +  ....  + \UF_{\gamma_m}(f_\lambda)$.
\end{corollary}

%% file: sections/50_experiments.tex
\section{Experiments}\label{sec:experiments}
\begin{figure*}[t]
    \centering
    \captionsetup[subfigure]{justification=centering}
    \begin{subfigure}{.32\textwidth}
        \centering
        \includegraphics[width=0.9\linewidth]{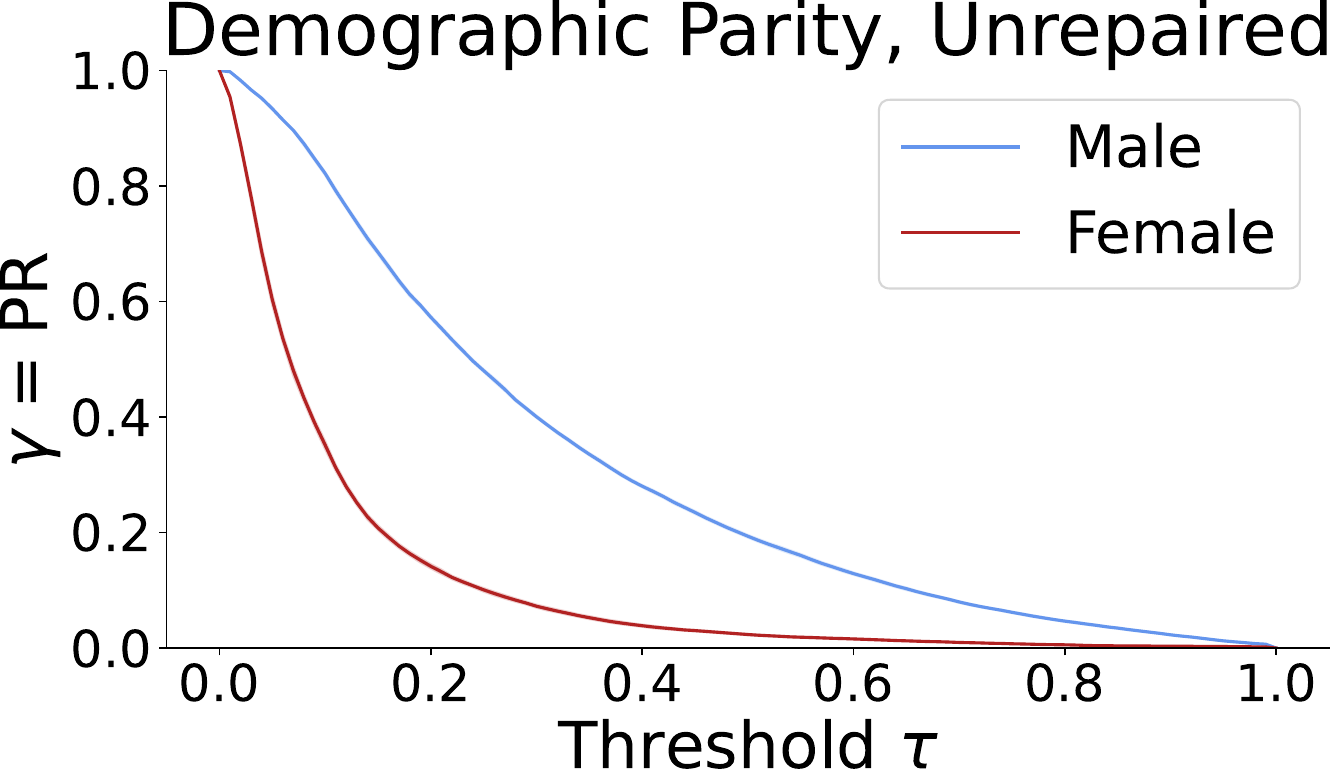}\\
        \includegraphics[width=0.9\linewidth]{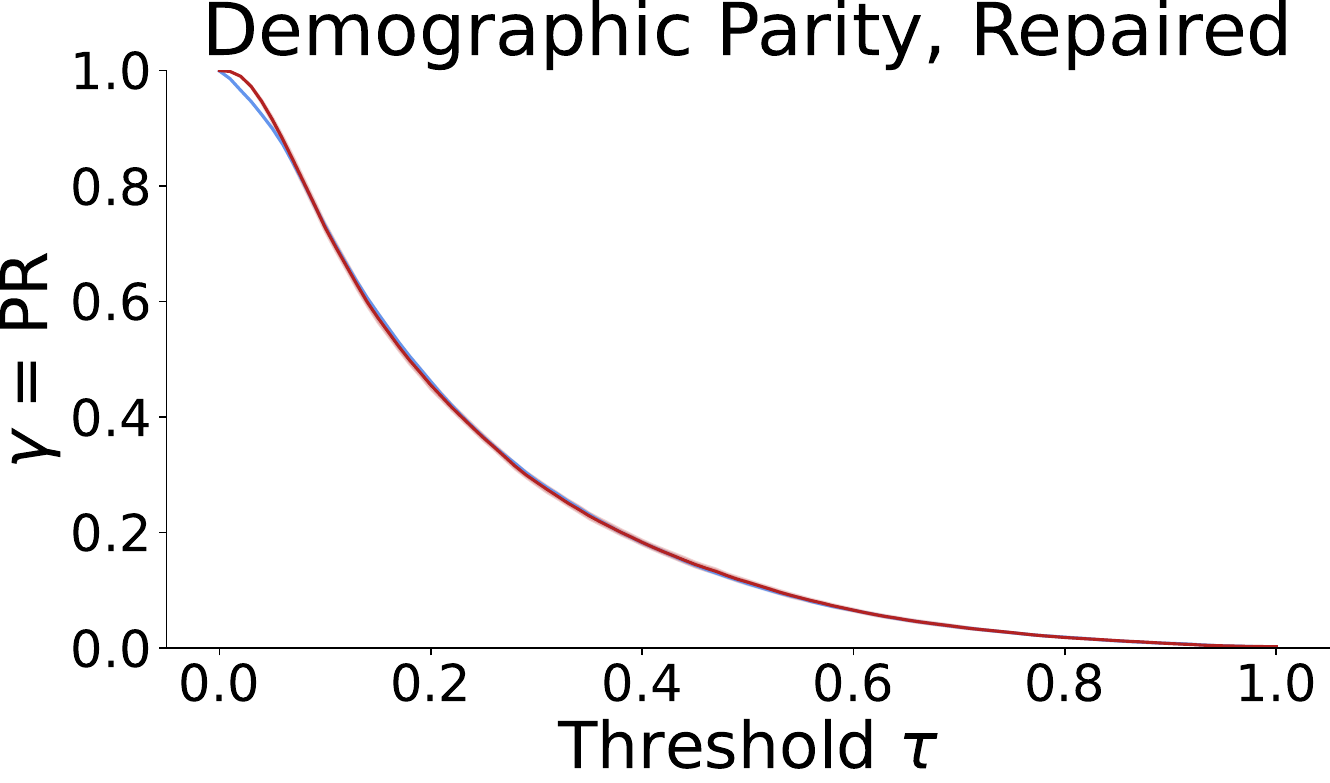}
        \subcaption{Unrepaired vs. Full Repair}
        \label{fig:lr-sdp}
    \end{subfigure}%
    \begin{subfigure}{0.64\textwidth}
        \centering
        \includegraphics[width=0.45\linewidth]{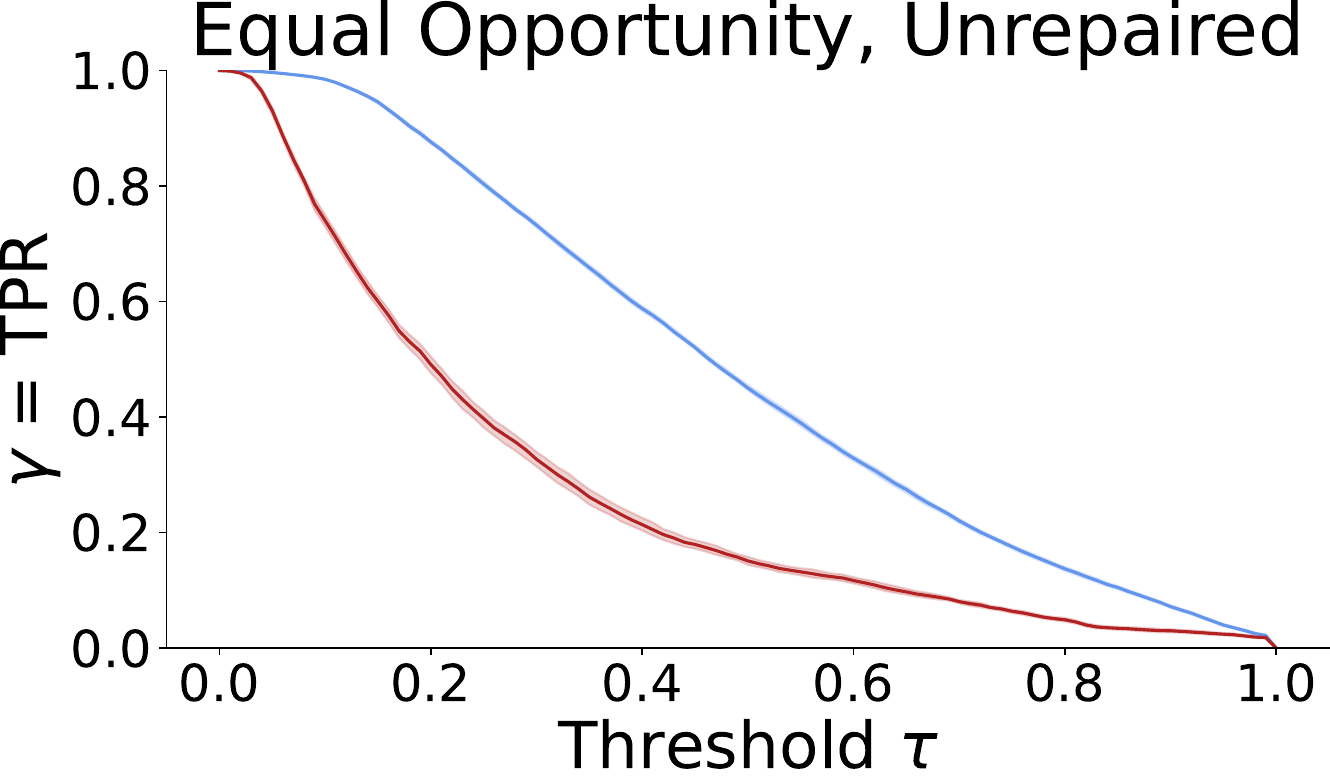}
        \includegraphics[width=0.45\linewidth]{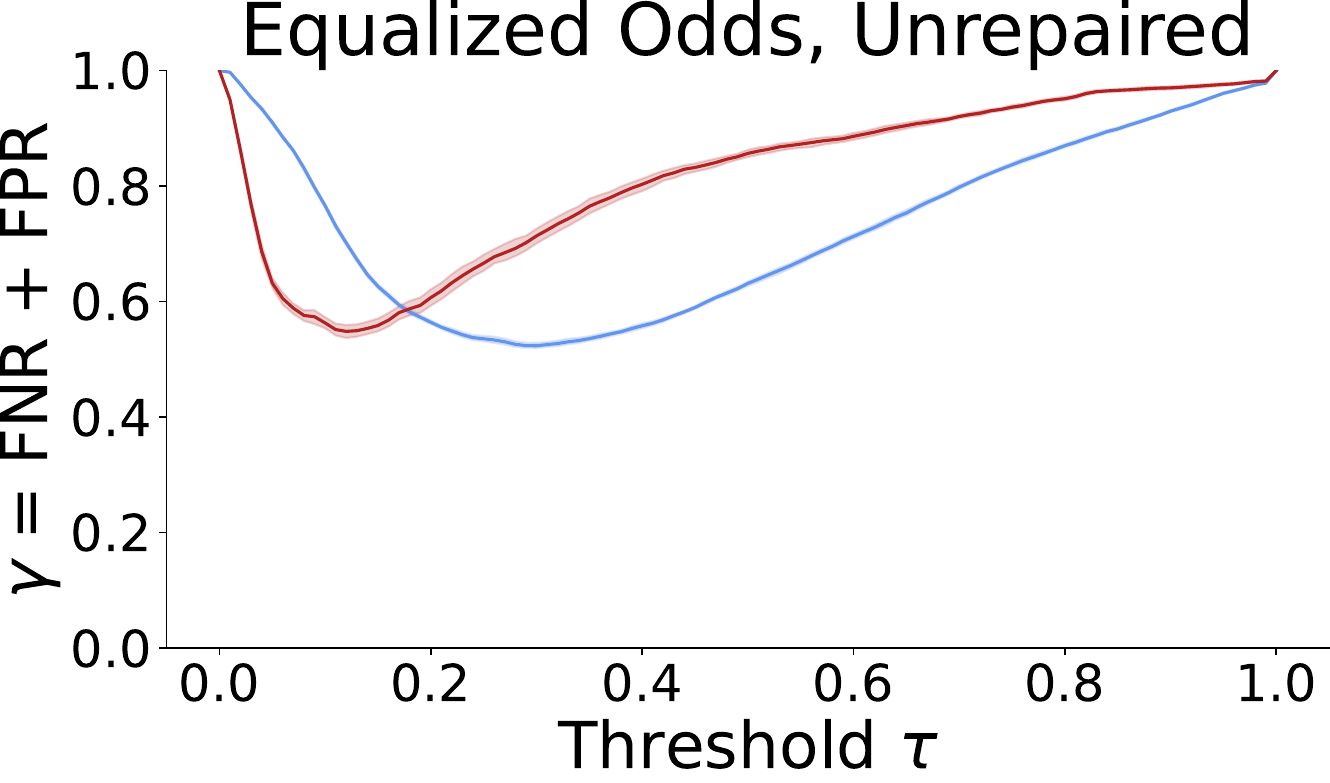}\\
        \includegraphics[width=0.45\linewidth]{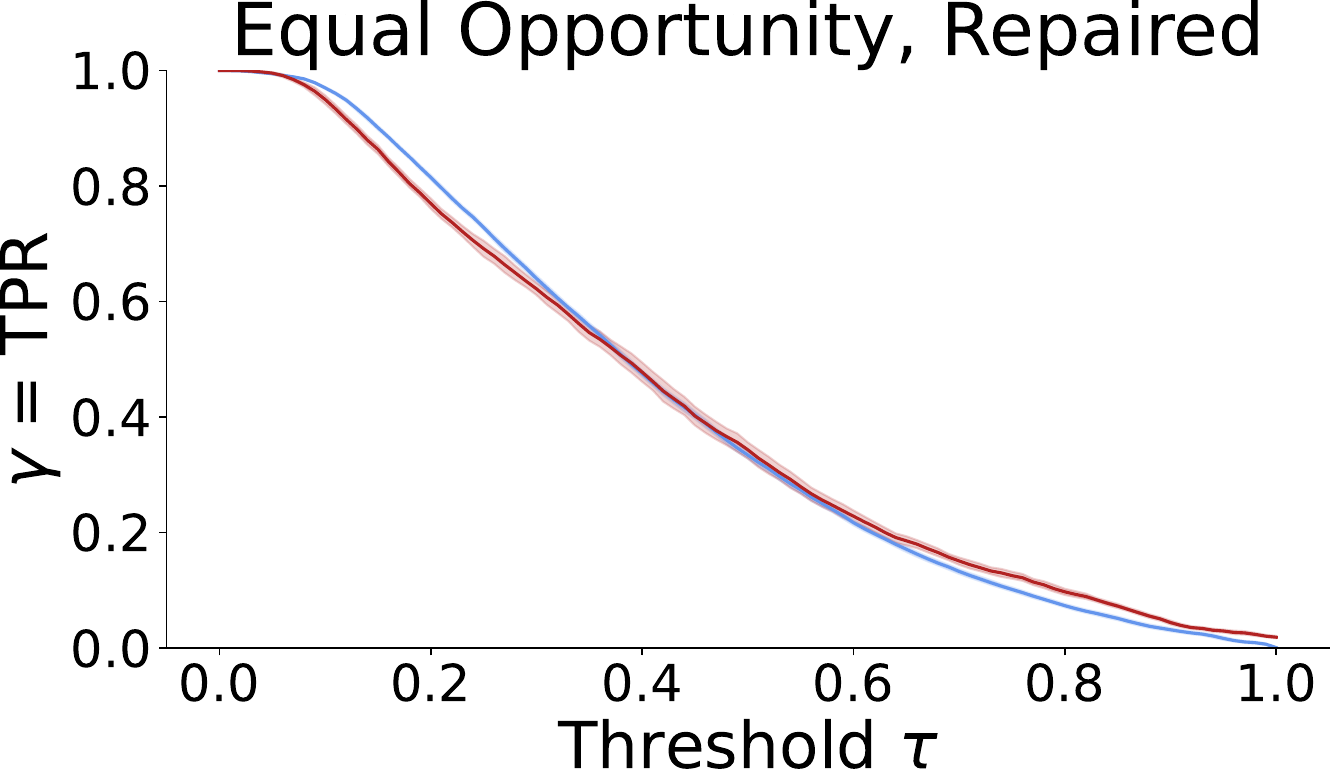}
        \includegraphics[width=0.45\linewidth]{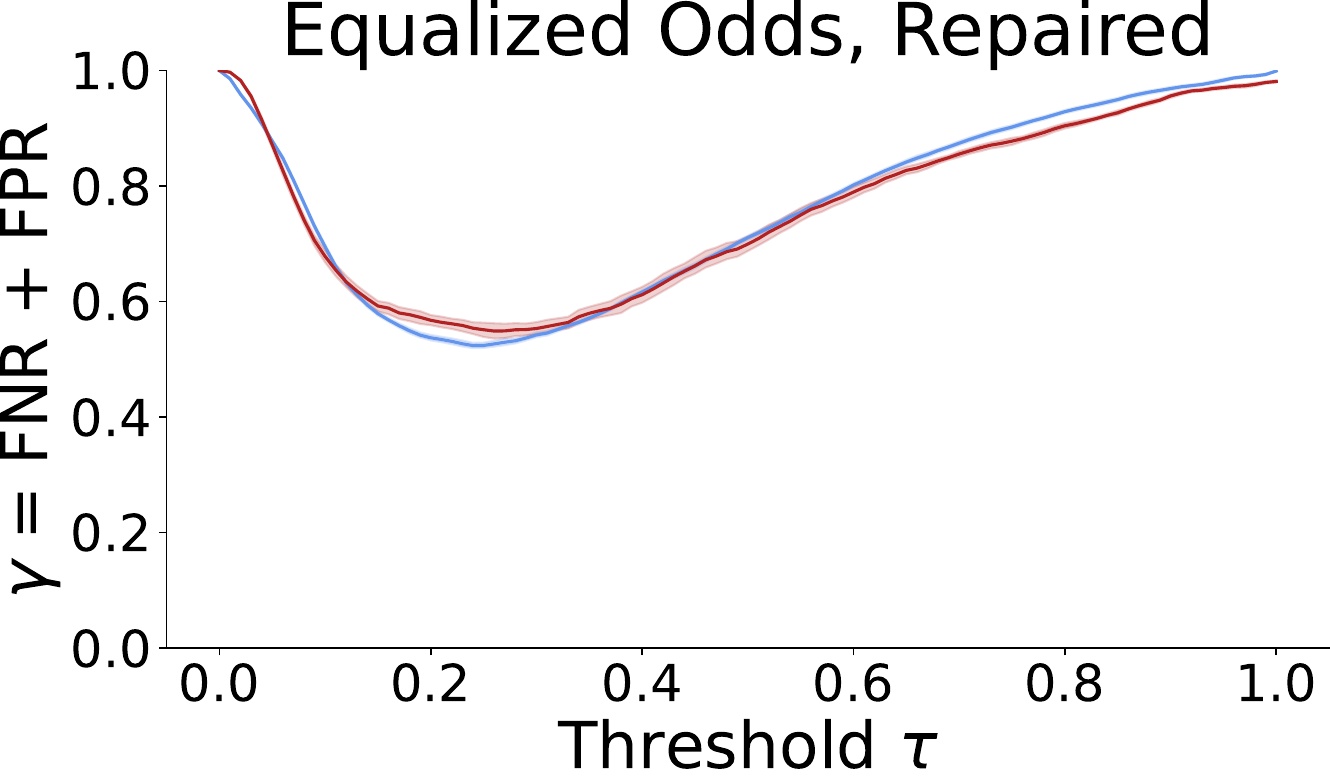}\\
        \subcaption{Unrepaired vs. Optimal Repair}
        \label{fig:lr-ours}
    \end{subfigure}
    \vspace{-.2cm}
       \caption{Performing geometric repair for $\gamma=\PR$ (left), \TPR (middle), $\EO$ (right)  for Logistic Regression trained on \texttt{Adult} Income-Race. The top row depicts the rates for unrepaired regressors and the bottom row for the repaired regressor.}

    \label{fig:lr}
    \vspace{-.2cm}
\end{figure*}

In this section, we present experiments that demonstrate the effectiveness of our proposed algorithms. We present two sets of results: 1)  Figure \ref{fig:lr} validates that Algorithm~\ref{alg:repair} achieves almost-exact distributional parity for Demographic Parity, Equal Opportunity, and Equalized Odds; 2)  Table \ref{table:baseline} shows that Algorithm~\ref{alg:repair} outperforms related methods in maximizing Distributional parity while preserving accuracy. \looseness=-1

\begin{table*}[]
\centering
\scriptsize
\caption{Comparison of Geometric Repair (\textbf{GR}) against included baselines (abbreviations described under \textbf{baselines}).  Results are averaged over ten trials, and the mean and standard deviation across all trials are reported for each metric.}
\label{table:baseline}
\begin{tabular}{cccccccc} \toprule
&    & \multicolumn{3}{c}{$\TPR$ Optimized}  & \multicolumn{3}{c}{$\EO$ Optimized}  \\ 
\multicolumn{1}{c}{}                                                                             & \multicolumn{1}{c}{}    & $\UF_\TPR$               & \multicolumn{1}{c}{Worst case}               & \multicolumn{1}{c}{AUC}             & $\UF_\EO$                & Worst case                                    & \multicolumn{1}{c}{AUC}             \\ \midrule
\multicolumn{1}{c|}{\multirow{4}{*}{\begin{tabular}[c]{@{}c@{}} Income-Race \\ (LR)\end{tabular}}}  & \multicolumn{1}{c|}{GR}  & \textbf{0.025 {$\pm$} 0.013} & \multicolumn{1}{c|}{\textbf{0.068 {$\pm$} 0.034}} & \multicolumn{1}{c|}{\textit{0.816 {$\pm$} 0.004}} & \textbf{0.021 {$\pm$} 0.007} & \multicolumn{1}{c|}{\textbf{0.055 {$\pm$} 0.02}}  & \multicolumn{1}{c}{\textit{0.816 {$\pm$} 0.005}} \\
\multicolumn{1}{c|}{}                                                                            & \multicolumn{1}{c|}{JIA} & 0.028 {$\pm$} 0.007          & \multicolumn{1}{c|}{0.129 {$\pm$} 0.021}          & \multicolumn{1}{c|}{0.8 {$\pm$} 0.005}   & 0.049 {$\pm$} 0.006          & \multicolumn{1}{c|}{0.094 {$\pm$} 0.014}          & \multicolumn{1}{c}{0.8 {$\pm$} 0.005}   \\
\multicolumn{1}{c|}{}                                                                            & \multicolumn{1}{c|}{FEL} & 0.042 {$\pm$} 0.039          & \multicolumn{1}{c|}{0.106 {$\pm$} 0.051}          & \multicolumn{1}{c|}{0.81 {$\pm$} 0.015}  & 0.103 {$\pm$} 0.039          & \multicolumn{1}{c|}{0.213 {$\pm$} 0.081}          & \multicolumn{1}{c}{0.811 {$\pm$} 0.014} \\
\multicolumn{1}{c|}{}                                                                            & \multicolumn{1}{c|}{OG}  & 0.219 {$\pm$} 0.01           & \multicolumn{1}{c|}{0.430 {$\pm$} 0.024}          & \multicolumn{1}{c|}{0.834 {$\pm$} 0.005} & 0.142 {$\pm$} 0.007          & \multicolumn{1}{c|}{0.299 {$\pm$} 0.016}          & \multicolumn{1}{c}{0.834 {$\pm$} 0.005} \\ \hline \hline

\multicolumn{1}{c|}{\multirow{4}{*}{\begin{tabular}[c]{@{}c@{}}Income-Sex \\ (SVM)\end{tabular}}} & \multicolumn{1}{c|}{GR}  & \textbf{0.014 {$\pm$} 0.005} & \multicolumn{1}{c|}{\textbf{0.042 {$\pm$} 0.015}} & \multicolumn{1}{c|}{\textit{0.882 {$\pm$} 0.004}} & 0.032 {$\pm$} 0.006          & \multicolumn{1}{c|}{0.079 {$\pm$} 0.015} & \multicolumn{1}{c}{\textit{0.882 {$\pm$} 0.004}} \\
\multicolumn{1}{c|}{}                                                                            & \multicolumn{1}{c|}{JIA} & 0.052 {$\pm$} 0.009          & \multicolumn{1}{c|}{0.104 {$\pm$} 0.013}          & \multicolumn{1}{c|}{0.878 {$\pm$} 0.004} & 0.047 {$\pm$} 0.006          & \multicolumn{1}{c|}{0.110 {$\pm$} 0.011}          & \multicolumn{1}{c}{0.878 {$\pm$} 0.004} \\
\multicolumn{1}{c|}{}                                                                            & \multicolumn{1}{c|}{FEL} & 0.014 {$\pm$} 0.007          & \multicolumn{1}{c|}{0.08 {$\pm$} 0.015}           & \multicolumn{1}{c|}{0.769 {$\pm$} 0.007} & \textbf{0.026 {$\pm$} 0.006} & \multicolumn{1}{c|}{0.081 {$\pm$} 0.013}          & \multicolumn{1}{c}{0.769 {$\pm$} 0.007} \\
\multicolumn{1}{c|}{}                                                                            & \multicolumn{1}{c|}{OG}  & 0.065 {$\pm$} 0.01           & \multicolumn{1}{c|}{0.114 {$\pm$} 0.015}          & \multicolumn{1}{c|}{0.884 {$\pm$} 0.003} & 0.035 {$\pm$} 0.008          & \multicolumn{1}{c|}{\textbf{0.077 {$\pm$} 0.013}}          & \multicolumn{1}{c}{0.884 {$\pm$} 0.003} \\ 
\bottomrule
\end{tabular}
\vspace{-.65cm}
\end{table*}

\vspace{-.2cm}
\subparagraph*{Datasets.} We use two datasets: \texttt{Adult} Income-Sex from the the UCI repository~\cite{dua2017uci}, and \texttt{Adult} Income-Race from the datasets produced in~\cite{ding2021retiring}. For both datasets, the task is to predict whether ($1$) or not ($0$) an individual's income exceeds \$50,000. 

\vspace{-.2cm}
\subparagraph*{Model Training.} To produce a model that we use in our experimentation, we implemented a Logistic Regression (\textbf{LR}) with $\ell_2$ regularization, and a Support Vector Machine (\textbf{SVM}) with a Radial Basis Function kernel. Both were implemented using scikit-learn with its default model parameters and optimizers~\cite{pedregosa2011scikit}.  

\vspace{-.2cm}
\subparagraph*{Metrics. }  We use the following measurements of model performance: (1) We approximate \textbf{Distributional parity $\UF_\gamma(\cdot)$} as per Step 2 of Algorithm \ref{alg:repair} using $m=100$ randomly sampled thresholds. We denote the Equalized Odds metric $\EO = \FNR + \FPR$, i.e., the misclassification rate(2) We measure accuracy using the \textbf{Area Under the Curve (AUC)} given that AUC averages model performance across all thresholds similar to $\UF_\gamma$. (3) \textbf{Worst Case} refers to the worst disparity of the regressor at any threshold for the chosen $\gamma$, i.e., $\max_{\tau \in \binaryrange}| \gamma_a(\tau) - \gamma_b(\tau)|$.  


\vspace{-.2cm}
\subparagraph*{Baselines. }  We use the following algorithms as baselines: (\textbf{OG})  The learned classifier with no additional processing. (\textbf{JIA}) Post-processing algorithm proposed by Jiang et al. \cite{jiang2020wasserstein} which processes the output of a regressor such that model output is independent of protected group (shown to be equal to satisfying $\UF_\PR = 0$, which is achieved by our method for $\lambda=1$). (\textbf{FEL})  Pre-Processing of model inputs from Feldman et al. \cite{feldman2015certifying} which seeks to reduce disparate impact across all thresholds.  
The "amount" of pre-processing is parametrized by a $\lambda$ similar to ours (just over inputs) -- we search for the optimal $\lambda$ for each metric we compare against.  We abbreviate geometric repair with (\textbf{GR}).



\vspace{-.2cm}
\subparagraph{Results.}  In Table \ref{table:baseline}, as denoted by the bolded cells in the $\UF_\gamma$ and \textit{Worst Case} columns, our method outperforms almost all baselines on both the \texttt{Adult} Income-Sex and \texttt{Adult} Income-Race tasks datasets, for both $\TPR$ and $\EO$. The one exception is for $\gamma=\EO$ on the Income-Sex task, however our method still attains a reduction in all-threshold disparity and preserves significant accuracy. For the AUC column, we italicize the cell that has AUC closest to that of the original regressor; for both metrics and datasets, our method was superior to the baselines in this aspect. We illustrate the effect of geometric repair at every threshold in Figure \ref{fig:lr}. For $\gamma=\PR$ (left) we show the \textit{full} correction $\lambda=1$. For $\gamma=\TPR$ (middle) we the computed optimal repair parameter $\lambdastar \approx 0.73 \pm 0.04$, and for $\gamma = \EO$ (right) we computed $\lambdastar \approx 0.75 \pm 0.03$.

%% file: sections/60_conclusion.tex
\vspace{-.2cm}
\section{Discussion and Related Work}\label{sec:conclusion}
\vspace{-.2cm}
In this work, we show that by interpolating between group-conditional score distributions we can achieve all-threshold fairness on fairness metrics like Equal Opportunity and Equalized Odds. To this end, we introduce Distributional parity to measure parity in a fairness metric at all thresholds, and provide a novel post-processing algorithm that 1) is theoretically grounded by our convexity result, and 2) performs well across benchmark datasets and tasks. 

A number of prior works have demonstrated how to achieve exact distributional parity in the special case when $\gamma = \PR$. Our work is most closely related to \cite{jiang2020wasserstein} who accomplish this using the $\Wass_{1}$ distance, in both in/post-processing settings.    
\cite{chzhen2020fairwasserstein, legouic2020projection} report a similar post-processing result to ours, deriving an optimal fair predictor (also limited to $\gamma=\PR$) in a regression setting and using $\Wass_2$ barycenters.  We build on these approaches by extending them to a broader class of fairness metrics and definitions. Our technique is based on the \textit{geometric repair} algorithm which was originally introduced by \cite{feldman2015certifying}  as a way to navigate the fairness-accuracy trade-off.  Geometric repair was also studied by \cite{gordaliza2019obtaining}. 
In the post-processing setting, the effect of geometric repair on classifier accuracy and $\gamma=\PR$ fairness was studied in \cite{chzhen2022minimax} -- we extend these 
to all $\gamma \in \Gamma$ by showing convexity on the set of regressors enumerated by geometric repair.

%% file: sections/appendix/appendix.tex
\newpage
\appendix
\input{sections/appendix/background.tex}

\input{sections/appendix/proofs.tex}

%% file: sections/appendix/background.tex
\onecolumn
\section{Additional Background on Optimal Transport}

In this section of the appendix, we present some additional background and theory from Optimal Transport.  These results are necessary to prove some of the results in the main paper body.

\subsection{Wasserstein Geodesics}
One key property of Wasserstein Barycenters that we exploit in this work, which is not refereed to explicitly in the main paper body, is that Wasserstein Barycenters under Geometric Repair form a curve in the space of probability measures called a constant speed geodesic.  

\begin{definition}[\cite{santambrogio2015optimal}, pg. 182] 
\label{def:geodesic}

Let $(X,d)$ be some metric space. A curve $\omega:\ospace \rightarrow X$ is a \emph{constant speed geodesic} between $\omega(0)$ and $\omega(1)$ if it satisfies 
\begin{align*}
    d(\omega(t), \omega(s)) = |{t - s}|d(\omega(0), \omega(1)) \quad \forall t,s \in \ospace 
\end{align*}
\end{definition}

The following result from \cite[Theorem 5.27]{santambrogio2015optimal} proves that a specific interpolation of an optimal transport plan forms a geodesic in the space of probability measures metricized by the Wasserstein Distance.\footnote{We can metricize $\Ppm{p}$ with $\Wass_p$ under \cite[Theorem 6.9]{Villani2008OptimalTO}} We also remind the reader that in the following expression, $\#$ denotes the pushforward operator on measures and $\id$ denotes the identity function. \footnote{In this section, we use a sub-scripted $\mu_t$ to denote a measure that is the result of some interpolation when clear from context; this subscript notation should not to be confused with the sub-scripts used on measures, e.g. $\mu_2$, in other places in the paper.}  

\begin{theorem}
\label{thm:geodesic_thm}
Suppose that $\Omega$ is convex, take $\mu, \nu \in \Ppm{p}(\Omega)$, and $\gamma \in \Gamma(\mu, \nu)$ an optimal transport plan for the cost $c(x,y) = |x-y|^p$ w/ ($p \geq 1$). Define $\pi_t: \Omega \times \Omega \rightarrow \Omega$ through $\pi_t(x,y) = (1-t)x + ty$. Then the curve $\mu_t(\pi_t)_{\#}\gamma$ is a constant speed geodesic connecting $\mu_0 = \mu$ to $\mu_1 = \nu$. 
In the particular case where $\mu$ is absolutely continuous then this curve is obtained as $((1-t)id + tT)_{\#\mu}$
\end{theorem}

For $p=2$, this special form of the interpolation between measures given in the above theorem is actually the exact same interpolation that is carried out by Wasserstein Barycenters.\footnote{This result is stated in  \cite{agueh2011barycenters} as the conclusion of Section 4 (see eq. 4.10) and in Section 6.2 of the same work}  
\begin{proposition}[\cite{agueh2011barycenters}]
\label{claim:mcann}
Let $\mu, \nu \in \Ppm{2}(\ospace)$ satisfy Assumption \ref{assumption:wass} then $\alpha$-weighted barycenters 
\begin{align*}
\mu_\alpha \leftarrow \argmin_{\in \Ppm{p}(\ospace)}  (1-\alpha)\Wass_2^2(\mu, \cdot) + \alpha\Wass_2^2(\nu, \cdot),  
\end{align*} can be equivalently computed $((1-t)id + tT)_{\#\mu}$ where $T$ is the transport plan that solves transport from $\mu \rightarrow \nu$. 
\end{proposition}

This means, under our mild assumptions, that barycenters both (a) follow the special form in Theorem \ref{thm:geodesic_thm} and (b) are constant speed geodesics.  We use this fact to show that the distance between $\lambda$ repaired measures $\mu_{\pa,\lambda}, \mu_{\protp, \lambda}$ can be written as a $1-\lambda$ weighted fraction of the Wasserstein distance of the unrepaired measures $\mu_\pa, \mu_\protp$. 
\begin{proposition}
\label{cor:geodesic}
Since $\mu_{\prot, \lambda}$ is a constant speed geodesic, the Wasserstein distance between repaired measures is proportional to the repair amount, i.e.,  $\Wass_1(\mu_{\pa, \lambda}, \mu_{\protp,\lambda}) = (1-\lambda)\Wass_1(\mu_\pa, \mu_\protp)$. 
\begin{proof}
Let $\mu_\pa, \mu_\protp \in \Ppm{2}$ and $T_a^b$ be the optimal transport plan from $\mu_\pa \rightarrow \mu_\protp$.  Suppose we parametrize the interpolation from $\mu_a$ to $\mu_b$ with a function $w:\binaryrange \rightarrow \Ppm{1}(\binaryrange)$ where $w(\alpha) = ((1-\alpha)\id + \alpha T_\pa^\protp)\#\mu_\pa$.  By Theorem \ref{thm:geodesic_thm}, this curve is a constant speed geodesic.  Now, consider the geometric repair score distributions $\mu_{\pa,\lambda}$ and $\mu_{\protp,\lambda}$.  We see from  Proposition \ref{cor:geodesic} that each distribution $\mu_{\prot, \lambda}$ is the result of the $\lambda$-weighted interpolation of $\mu_\prot$ to the barycenter $\mustar$.  These barycenters can alternatively computed by interpolating from $\mu_\pa \rightarrow \mu_\protp$, i.e.,  
\begin{align*}
    \mu_{\pa, \lambda} &=  ((1 - \lambda \rho_\protp) \id + \lambda \rho_\protp T_\pa^\protp)\#\mu_\pa \\ 
     \mu_{\protp, \lambda} &= (\lambda \rho_\pa \id + (1-\lambda \rho_\pa) T_\pa^\protp)\#\mu_\pa . 
\end{align*}
From this, a reparametrization of the above interpolation under geometric using $w(\cdot)$ yields,   
\begin{align*}
    \mu_{\pa, \lambda} = w(\lambda \rho_\protp) \quad \text{and} \quad \mu_{\protp, \lambda} = w(1- \lambda \rho_\pa). 
\end{align*}

Then, the corollary follows from the definition of constant speed geodesics in Definition \ref{def:geodesic}, i.e., 
\begin{align}
    \Wass_{1}(\mu_{\pa, \lambda}, \mu_{\protp,\lambda}) =& ~\Wass_{1}(w(\lambda \rho_\protp), w(1-\lambda \rho_\pa)) \\ =& ~|\lambda\rho_\protp - (1 - \lambda\rho_\pa)|\Wass_{1}(\mu_\pa, \mu_\protp) \\ =&   
    ~|\lambda(\underbrace{\rho_\pa + \rho_\protp}_{~=1~\text{by Def}}) - 1|\Wass_{1}(\mu_\pa, \mu_\protp) \\ =& ~(1-\lambda)\Wass_{1}(\mu_\pa, \mu_\protp)
\end{align}
\end{proof}
\end{proposition}

The last additional result we'll need to aid our effort to prove Theorem \ref{thm:convex}, is the following Lemma. Please note this lemma differs from the above corollary due to the specific optimal transport problems being solved. In the above, we are consider a parametrization of $\mu_{\prot, \lambda}$ along the interpolation from $\mu_\pa \rightarrow \mu_\protp$.  In the below result we consider the interpolation of repaired distributions to their barycenter, i.e., $\mu_{\pa,\lambda} \rightarrow \mustar$. 
\begin{lemma}
\label{lemma: subsitution_help}
Let $\mu_\pa, \mu_\protp \in \Ppm{2}(\ospace)$ satisfy Assumption \ref{assumption:wass} and let $\mu_{\pa,\lambda}$ be the $\lambda$-barycenter of $\mu_a$ and $\mu_*$, and let $\mu_{\protp,\lambda}$ be the $\lambda$-barycenter of $\mu_\protp$ and $\mu_*$ then 
\begin{align*}
    \mu_{\pa,\lambda} &= \mu_{\protp, \frac{1-\rho_\pa\lambda}{1-\rho_\pa}} \\ 
    \mu_{\protp, \lambda} &= \mu_{\pa, \frac{1-\lambda}{\rho_\pa} + \lambda}
\end{align*}
\begin{proof}
Let $\mustar$ be the $\rho_\protp$ barycenter of $\mu_\pa, \mu_\protp$.  It is easy to show from their definitions that $\mu_{\pa,\lambda_1} = \mu_{\lambda_1(1-\rho_\pa)}$ and  $\mu_{\protp,\lambda_2} = \mu_{1-\lambda_2 \rho_\pa}$ (Figure \ref{fig:repairtimeline} provides a nice illustration of this fact).  To prove the Lemma, we let $\lambda_1(1-\rho_\pa) = 1-\lambda_2 \rho_\pa$. Solving for $\lambda_1$, yields the proposition, i.e., $\lambda_1 = \frac{1- \lambda_2 \rho_\pa}{\rho_\protp}$ and therefore $\mu_{\pa,\lambda_1} = \mu_{\protp, \frac{1-\rho_\pa\lambda_2}{\rho_\protp}}$. Letting $\lambda_1 = \lambda_2$, such that both $\mu_{a}, \mu_\protp$ are controlled by the same repair parameter yields the first equality. 
 Solving for $\lambda_2$ and making the same substitution ($\lambda_2 = \lambda_1)$ yields the second equality. 
\end{proof}
\end{lemma}

\subsection{The Relationship Between Fair Risk Minimization and Barycenters}

In this subsection we give an additional result relating the lowest risk $\gamma=\PR$ regressor to the distance of that regressors groupwise score distributions, to their barycenter.  

\begin{lemma}
\minfairrisklemma
\end{lemma}

\begin{proof}
Suppose $h$ is the regressor which minimizes the l.h.s. and let $\mu_h = Law(h(\Xrv, \Protrv))$.  We can re-write 
\begin{align*}
 \sum_{\prot \in \prot} p_\prot \Wass_1(\mu_g, \mu_h) &= \sum_{\prot \in \prot} p_\prot \min_{T \in \mc T_\prot^h} \int_{\ospace} | x - T(x)|d\mu_\prot \\
 &= \sum_{\prot \in \prot} p_\prot \min_{T \in \mc T_\prot^h} \int_{X} |f(\Xrv,g) - T(f(\Xrv,g))| d \mu_{X \rvert \prot} 
 \end{align*}
Let $T_{\prot}^{\hhat} = F_{\mu_{\hhat}} \circ F^{-1}_{\mu_\prot}$ be the optimal transport maps which minimize the above, and let $\hat{h}(x,\prot) = T_\prot^h(f(x,\prot))$.  We can continue 
 \begin{align*}
  \sum_{\prot \in \prot} p_\prot \min_{T \in \mc T_\prot^h} \int_{X} |f(\Xrv,g) - \hhat(\Xrv,g))| d \mu_{X \rvert \prot} &=  \Expect_{\prot \sim \Protrv }\left[ \Expect_{\Xrv}  [|f(\Xrv,g) - \hhat(\Xrv,g)| ]\rvert \Protrv =\prot \right] \\ 
 &= \Expect_{\Xrv, \Protrv} \left [|f(\Xrv,\Protrv) -  \hhat(\Xrv,\Protrv)| \right ].  
\end{align*}
From the above equalities we've shown,
\begin{align}
\label{step: lemmaa1step0}
    \sum_{\prot \in \prot} p_\prot \Wass_1(\mu_a, \mu_h) = \Expect_{\Xrv, \Protrv} \left [|f(\Xrv,\Protrv) -  \hhat(\Xrv,\Protrv)| \right ].  
\end{align}
and by the presumed optimality of $h$ it follows,
\begin{align}
\label{step:lemmaa1step1}
   \Expect_{\Xrv, \Protrv} \left [|f(\Xrv,\Protrv) -  \hhat(\Xrv,\Protrv)| \right ] \geq \Expect_{\Xrv, \Protrv} \left [|f(\Xrv,\Protrv) -  \h(\Xrv,\Protrv)| \right ].
\end{align}
On the other hand suppose $T_\prot^h$ is an optimal transport plan such that $h(x,\prot) = T_\prot^h(f(x,\prot))$ then, by the optimality of $T_*^{\hhat}$ it follows 
\begin{align*}
\sum_{\prot \in \prot} p_\prot \int_{X} |f(\Xrv,g) - T_\prot^{\hhat}(f(\Xrv,g))| d \mu_{X \rvert \prot} \leq  \sum_{\prot \in \prot} p_\prot \int_{X} |f(\Xrv,g) - T_\prot^h(f(\Xrv,g))| d \mu_{X \rvert \prot}. 
\end{align*}
Using similar properties as the above derivations we can re-write this relationship as 
\begin{align}
\label{step:lemmaa1step2}
    \Expect_{\Xrv, \Protrv} \left [|f(\Xrv,\Protrv) -  \hhat(\Xrv,\Protrv)| \right ] \leq \Expect_{\Xrv, \Protrv} \left [|f(\Xrv,\Protrv) -  \h(\Xrv,\Protrv)| \right ].
\end{align}
Therefore by Steps (\ref{step:lemmaa1step1}) and (\ref{step:lemmaa1step2}) we have 
\begin{align*}
     \Expect_{\Xrv, \Protrv} \left [|f(\Xrv,\Protrv) -  \hhat(\Xrv,\Protrv)| \right ] = \Expect_{\Xrv, \Protrv} \left [|f(\Xrv,\Protrv) -  \h(\Xrv,\Protrv)| \right ], 
\end{align*}
and combining Step \ref{step: lemmaa1step0} with the above concludes 
\begin{align}
\label{step: lemmaa1step3}
     \min_{\nu \in \Ppm{1}(\ospace)} \sum_{\prot \in \prot} p_\prot \Wass_1(\mu_a, \nu) \leq \Expect_{\Xrv, \Protrv} \left [|f(\Xrv,\Protrv) -  \h(\Xrv,\Protrv)| \right ],
\end{align}
where $\UF_\PR(h) = 0 $ by assumption. To prove the other direction, now let 
\begin{align*}
    \bar{\nu} \leftarrow \argmin_{\nu \in \Ppm{1}(\ospace)} \sum_{\prot \in \prot} p_\prot \Wass_1(\mu_a, \nu)
\end{align*}
and $T_\prot^{\bar{\nu}}$ be the optimal transport maps from $\mu_\prot \rightarrow \bar{\nu}$ and $\bar{h}(x,\prot) = T_\prot^{\bar{\nu}}(f(x,\prot))$.  Now, if we consider  
\begin{align*}
    \sum_{\prot \in \prot} p_\prot \Wass_1(\mu_a, \bar{\nu}) =  \Expect_{\Xrv, \Protrv} \left [|f(\Xrv,\Protrv) -  \bar{h}(\Xrv,\Protrv)| \right ]
\end{align*}
then we can easily conclude by the assumed optimality of $h$ that, 
\begin{align}
\label{step: lemmaa1step4}
   \min_{\nu \in \Ppm{1}(\ospace)} \sum_{\prot \in \prot} p_\prot \Wass_1(\mu_a, \nu) \geq \Expect_{\Xrv, \Protrv} \left [|f(\Xrv,\Protrv) -  \h(\Xrv,\Protrv)| \right ].
\end{align}
Finally, recalling that $\bar{h}$ satisfies $\UF_\PR(\bar{h})=0$ since $\Bar{h}$ is a Barycenter (Corollary \ref{cor:wass_sdp}). Combining Steps \ref{step: lemmaa1step3} and \ref{step: lemmaa1step4} to yield the proof. 
\end{proof}

%% file: sections/appendix/proofs.tex
\section{Proofs}
\subsection{Proof of Proposition \ref{cor:wass_sdp}}
\paragraph{Proposition \ref{cor:wass_sdp}. }
\wassspdcorollary
\begin{proof}
Let $\mu_\pa, \mu_\protp$ be the groupwise score distributions of some regressor $f$. Since $W_p$ is a metric on $\Ppm{p}(\ospace)$ (according to Proposition 2.3 in \cite{ComputationalOptimalTransport}) if $\Wass_2(\mu_\pa,\mu_\protp)$ then $\mu_\pa = \mu_\protp$. Similarly, by the same property we know that $\Wass_2(\mu_\pa,\mu_\protp) = \Wass_1(\mu_\pa,\mu_\protp) = 0$.  Showing that $\Wass_1(\mu_\pa,\mu_\protp) = \UF_\PR(f)$ completes the proof.  
To show this equality, recall by definition that  
\begin{align}
\gamma_{\prot}(\tau) 
&= \Pr[f(\Xrv,\Protrv) \geq \tau |  \Protrv=\prot] \\ 
&= 1 - \Pr[f(\Xrv,\Protrv) \leq \tau | \Protrv=\prot] \\ 
&= 1 - F_{\prot}(\tau)  
\end{align}
Plugging this into the expression for $\UF_\PR$  
\begin{align}
\UF_{\PR}(f) = \Expect_{\tau \in U(\ospace)}| \gamma_{\pa}(\tau) - \gamma_{\protp}(\tau)| 
&=  \int_{\ospace}| \gamma_{\pa}(\tau) - \gamma_{\protp}(\tau)|^{p}d\tau \\ 
&=  \int_{\ospace}| (1 - F_{\pa}(\tau)) - (1 - F_{\protp}(\tau)) |d\tau \\ 
&= \int_{\ospace}| F_{\pa}(\tau) - F_{\protp}(\tau) |d\tau \\
&= \int_{0}^{1}| F^{-1}_{\pa}(t) - F^{-1}_{\protp}(t) |dt = \Wass_{1}(\mu_{\pa}, \mu_{\protp})
\end{align} 
where the second to last equality was proven in Lemma 6 from \cite{jiang2020wasserstein}. 
\end{proof}

\subsection{Proof of Lemma \ref{lemma:opt_geometric_repair}}
\paragraph{Lemma \ref{lemma:opt_geometric_repair}. }
\geometricrepairfairoptimal
\begin{proof}
By Definition of $\fsdp$ we know that  $\mc R(\fsdp)$ is 
\begin{align*}
        \min_{\fhat \in \mc F} \mc  R(\fhat)  \quad \suchthat \quad \UF_{\PR}(\fhat) = 0. 
\end{align*}
It follows that 
\begin{align}
\label{step:paretotheoremstep0}
    \lambda(\min_{\fhat \in \mc F} \mc  R(\fhat)) = \min_{\fhat \in \mc F} \mc  \lambda R(\fhat) = \lambda\mc R(\fsdp).
\end{align}
By definition of $f_\lambda$ it is straightforward to show that $\mc R(f_\lambda) = \lambda \mc R(\fsdp)$.  Under Proposition \ref{cor:geodesic}, it is straightforward to show that $\UF_\PR(f_\lambda) = (1-\lambda)\UF_\PR(f)$.  Combining these two facts proves the result. 
\end{proof}

\subsection{Proof of Corollary \ref{cor:barycenter_transport_plan}}
\paragraph*{Corollary \ref{cor:barycenter_transport_plan}. }
Let $\mustar$ be the $\rho_\protp$-weighted barycenter of $\mu_\pa,\mu_\protp$ then the transport plan from $\mu_\pa \rightarrow \mustar$ (wlog) is computed
 \begin{align*}
T_\pa^*(\omega) =  (\rho_\pa F_{\mu_\pa} + \rho_\protp F_{\mu_\protp}) \circ F_{\mu_\pa}^{-1}(\omega) 
\end{align*}
\begin{proof}
Observe that by Theorem \ref{thm:geodesic_thm} we can express barycenter from $\mu_\pa$ to $\mustar$ (wlog) 
\begin{align*}
    \mustar = (\rho_\pa\id + \protp T_\pa^\protp)\#\mu_\pa &= (\rho_\pa F^{-1}_{\mu_\pa}\circ F_{\mu_\pa} + \rho_\protp F^{-1}_{\mu_\protp} \circ F_{\mu_\pa})\#\mu_\pa 
\end{align*}
The second equality follows from Remark \ref{rem:transport_plan}.  From this expression, we can define $T_\pa^* = (\rho_\pa F^{-1}_{\mu_\pa}\circ  + \rho_\protp F^{-1}_{\mu_\protp} )\circ F_{\mu_\pa}$ as the function which computes the transport from $\mu_\pa \rightarrow \mustar$.  
\end{proof}

\subsection{Proof of Proposition \ref{prop:lambda_relxation}
}
\paragraph{Proposition \ref{prop:lambda_relxation}
. }
For any $\lambda \in \binaryrange$, a repaired regressor $f_\lambda$ satisfies the following 
\begin{align*}
R(f_\lambda) = \lambda R(\fsdp) \quad\text{and}\quad
\UF_\PR(f_{\lambda})  = (1-\lambda)\UF_\PR(f)
\end{align*}
\begin{proof}
The first equality follows from the definition of $R$ and linearity of expectation. It is easy to show that 
\begin{align*}
    R(f_\lambda) &= R((1-\lambda)f + \lambda \fsdp) \\ &= (1-\lambda)R(f) + \lambda R(\fsdp) = \lambda R(\fsdp)
\end{align*}
where the last equality follows by noting that $R(f)=0$ by definition.  The proof that $\UF_\PR(f_{\lambda})  = (1-\lambda)\UF_\PR(f)$ follows from Proposition \ref{cor:geodesic}. 
\end{proof}

\subsection{Proof of Theorem~\ref{thm:convex}.} 
In the proof of Theorem \ref{thm:convex}, we make use of the fact that this transport plans are bijective, under Assumption \ref{assumption:wass}. In order to show that these plans are bijective we show that they are strictly monotone via the following Remark.  
\begin{remark}[\cite{santambrogio2015optimal}, p. 55]
\label{rem:continuous_monotone_T}
For two measures $\mu, \nu \in \Ppm{p}(\ospace)$, if $\nu$ is non-atomic, then the transport plan $T$ from $\mu \rightarrow \nu$ is strictly monotone on a closed domain like $\ospace$.
\end{remark}
It is well known that strictly monotone functions on a closed domain are bijective, and therefore we claim bijectivity as a corollary of the above result. 
\begin{corollary}
\label{lemma:bijection}
A transport plan $T$ that is strictly monotone, on a closed domain, is also bijective.  
\end{corollary}

Now we begin the proof of Theorem \ref{thm:geodesic_thm}. 
\label{appendix:pf:convexity}
\paragraph{Theorem \ref{thm:convex}. }\convexityresult
\begin{proof}
Let $\gamma \in \Gamma$. To prove convexity, we show that  $\frac{d^2}{d\lambda^2}\UF_\gamma(f_\lambda)$ is non-negative everywhere. First, we remind readers the definition of $\UF_\gamma(f_\lambda)$ (distributional parity): 
\begin{align*}
     \mathcal{U}_\gamma(f_\lambda) \triangleq \Expect_{\tau \sim U(\ospace)}| \gamma_{\pa}(\tau) - \gamma_{\protp}(\tau) |. 
\end{align*}
where $\gamma_\prot$ is a fairness metric on the score distributions of $f_\lambda$ for group $\prot \in \Prot$. 


Recall the definition of $\gamma_\prot(\tau; f_\lambda)$
\begin{align*}
    \gamma_\prot(\tau; f_\lambda) = \Pr[f_\lambda(\Xrv, \Protrv) \geq \tau | \Protrv = \prot]. 
\end{align*}
by Proposition \ref{prop:repaired_barycenters} we know that $\mu_{\prot, \lambda}$ is the score distribution associated with $f_\lambda(\cdot,\prot)$ and so we re-write this expression as a conditional expectation  
\begin{align}
\label{eq:metric_functional}
    \Pr[f_\lambda(\Xrv, \Protrv) \geq \tau | \Protrv = \prot] = \int_{\ospace} \mathbb{1}_{[\tau, 1]} d\mu_{\prot, \lambda}
\end{align}

In order to take this derivative, we need to invoke several change of variables to convert this Lebesgue integral to a Riemann integral.  We'll proceed for $\pa \in \Prot$ without loss of generality.  Also note for brevity, we present the proof for $\mu_{\prot,\lambda}$,i.e., the measure associated with $\gamma=\PR$.  Similarly, if we condition the l.h.s. of Eq. \ref{eq:metric_functional} on $\Yrv$,  our results follow similarly for corresponding probability measures associated with this conditional probability, .e.g, we would let $\mu_{\prot | \Yrv,\lambda}$ be the measure associated with the conditional probability $\Pr[f_\lambda(\Xrv, \Protrv)| \Protrv=\prot, \Yrv \geq \tau]$ for which setting $\Yrv$ computes $\TPR$ and $\FPR$ respectively.

Following Claim \ref{claim:mcann} can re-write $\mu_{\pa,\lambda} \coloneqq ((1-\lambda)id + \lambda T_\pa^*) \# \mu_\pa$.  For notational ease, define $\pi_{\pa,\lambda} \coloneqq (1-\lambda)id + \lambda T_\pa^*$.  Using these substitutions, we have that $\mu_{\pa, \lambda} = (\pi_{\pa, \lambda}) \# \mu_\pa$, so $\gamma_\pa$ can be equivalently written  
\begin{align*}
    \gamma_\pa(\tau) = \int_{\ospace} \mathbb{1}_{[\tau,1]} d(\pi_{\pa,\lambda}\#\mu_\pa). 
\end{align*}

By definition of the push-forward operator 
\begin{align*}
    \int_{\ospace} \mathbb{1}_{[\tau,1]} d(\pi_{\pa,\lambda}\#\mu_\pa) = \int_{\pi_{\pa,\lambda}^{-1}(\ospace)} \mathbb{1}_{[\tau,1]}(\pi_{\pa,\lambda}) d\mu_\pa =  \int_{\ospace} \mathbb{1}_{[\tau,1]}(\pi_{\pa,\lambda}) d\mu_\pa. 
\end{align*}
We note that the domain of integration is unchanged in the last equality because $\pi$ is a bijective mapping from $\ospace \rightarrow \ospace$ by Corollary \ref{lemma:bijection}, and so  $\pi_{\prot, \lambda}^{-1}(\ospace) = \ospace$.

For the last change of variables, Let $\ell$ be the Lebesgue measure. By Assumption \ref{assumption:wass} $\mu_a$ is absolutely continuous with respect to $\ell$ meaning that by the Radon Nikodym-Theorem
\begin{align*}
     \int_{\ospace} \mathbb{1}_{[\tau,1]}(\pi_{\prot,\lambda}) d\mu_\pa = \int_{\ospace} \sigma_{\pa}\mathbb{1}_{[\tau,1]}(\pi_{\prot,\lambda}) d\ell
\end{align*}
where $\sigma_{\pa}$ is the Radon-Nikodym Derivative, i.e., the probability density function associated with $\mu_\pa$.

We'll also need to define $\gamma_\protp$ similarly.  To do this we invoke Lemma \ref{lemma: subsitution_help} which yields that $\mu_{\protp,\lambda} = \mu_{\pa, \frac{1-\lambda}{\rho_\pa} + \lambda}$. Using this substitution we get 
\begin{align*}
    \gamma_\protp(\tau) = \int_{\ospace} \rho_{\mu_\pa}\mathbb{1}_{[\tau,1]}(\pi_{\protp, \frac{1-\lambda}{\rho_\pa} + \lambda}) d\ell 
\end{align*}
Next, let $h_{\pa,\tau}(\lambda)$ be the mapping $\lambda \mapsto \mathbb{1}_{[\tau,1]}(\mu_{\pa,\lambda})$ and $h_{\protp,\tau}(\lambda)$ be $\lambda \mapsto \mathbb{1}_{[\tau,1]}(\mu_{\pa, \frac{1-\lambda}{\rho_\pa} + \lambda})$.  Taking the first derivative of this difference, we get 
\begin{align*}
    \partialdd{}{\lambda} [h_{\pa,\tau}(\lambda) - h_{\protp,\tau}(
    \lambda)] &= \\   
    \partialdd{}{\lambda} \int_{\ospace} \rho_{\mu_\pa}\cdot [\mathbb{1}_{[\tau,1]}(\pi_{\pa,\lambda}) - \mathbb{1}_{[\tau,1]}(\pi_{\protp,\frac{1-\lambda}{\rho_\pa} + \lambda})] d\ell &= \\ \int_{\ospace} \rho_{\mu_\pa}\cdot\left[\partialdd{}{\lambda}\left(\mathbb{1}_{[\tau,1]}(\pi_{\pa,\lambda}) - \mathbb{1}_{[\tau,1]}(\pi_{\protp,\frac{1-\lambda}{\rho_\pa} + \lambda})\right)\right] d\ell   
\end{align*}
where the second equality follows from Leibniz Rule.  To finish the derivative, we remind the reader that the derivative of $\frac{d}{d\lambda}\mathbb{1}_{[\tau,1]}(\pi_{\prot,\lambda})$ is the Dirac delta function $\delta(\pi_{\prot,\lambda} - \tau)$.  It follows that
\begin{align*}
\int_{\ospace} \rho_{\mu_\pa}\cdot\left[\partialdd{}{\lambda}\left(\mathbb{1}_{[\tau,1]}(\pi_{\pa,\lambda}) - \mathbb{1}_{[\tau,1]}(\pi_{\protp,\frac{1-\lambda}{\rho_\pa} + \lambda})\right)\right] d\ell &= 
\int_{\ospace} \rho_{\mu_\pa} \cdot \biggr[T_\pa^*(\delta(\pi_{\pa,\lambda}-\tau)) +  \left(\frac{1-\rho_\pa }{\rho_\pa}\right)\delta(\pi_{\protp,\frac{1-\lambda}{\rho_\pa} + \lambda}-\tau)T_\protp^* \\
&- \left(\frac{\rho_\pa-1}{\rho_\pa}\right)\delta(\pi_{\protp,\frac{1-\lambda}{\rho_\pa} + \lambda}-\tau)\id - \delta(\pi_{\pa,\lambda}-\tau)\id)\biggr]
\end{align*}

and by definition of $\delta$ of the delta function, we at last obtain
\begin{align*}
     \int_{\ospace} \rho_{\mu_\pa} \cdot \left[T_\pa^*(\delta(\pi_{\pa,\lambda}-\tau)) +  \left(\frac{1-\rho_\pa }{\rho_\pa}\right)\delta(\pi_{\protp,\frac{1-\lambda}{\rho_\pa} + \lambda}-\tau)T_\protp^* - \left(\frac{\rho_\pa-1}{\rho_\pa}\right)\delta(\pi_{\protp,\frac{1-\lambda}{\rho_\pa} + \lambda}-\tau)\id - \delta(\pi_{\pa,\lambda}-\tau)\id)\right]  \\ = \left[T_\pa^* +  \left(\frac{1-\rho_\pa }{\rho_\pa}\right)T_\protp^* - \left(\frac{\rho_\pa-1}{\rho_\pa}\right)\id - \id\right] \circ \tau .
\end{align*}
To summarize, we have just shown that 
\begin{align*}
\partialdd{}{\lambda} [h_{\pa,\tau}(\lambda) - h_{\protp,\tau}(\lambda)] = \left[T_\pa^* +  \left(\frac{1-\rho_\pa }{\rho_\pa}\right)T_\protp^* - \left(\frac{\rho_\pa-1}{\rho_\pa}\right)\id - \id\right] \circ \tau. 
\end{align*} 
To prove convexity we must also compute the second derivative of the above.  Since the above does not depend on $\lambda$,  taking another derivative yields  
\begin{align}
\label{eq:secondderiv}
    \frac{d^2}{d\lambda^2} [h_{\pa,\tau}(\lambda) - h_{\protp,\tau}(\lambda)] = 0.
\end{align} Now, to prove the convexity of $\UF_\gamma(\f_\lambda)$ we take the second derivative of the absolute value of this difference, i.e., 
\begin{align}
\label{eq:convexity_inside_derivative}
   \frac{d}{d^2\lambda} |h_{\pa,\tau} - h_{\protp,\tau}| &= 
   \text{sign}(h_{\pa,\tau} - h_{\protp,\tau})\underbrace{\frac{d^2}{d\lambda^2} [h_{\pa,\tau}(\lambda) - h_{\protp,\tau}(\lambda)]}_{=~0} \\ &+ 2 \underbrace{\delta(h_{\pa,\tau} - h_{\protp,\tau})}_{\simeq ~0~\text{or}~1}\underbrace{(\partialdd{}{\lambda} [h_{\pa,\tau}(\lambda) - h_{\protp,\tau}(\lambda)])^2}_{\geq 0}. 
\end{align}
The first term on the r.h.s., we've already shown is zero, and the second term is also non-negative.  Another application of Leibniz' Rule allows that 
\[ 
      \frac{d}{d^2\lambda}\underbrace{\Expect_{\tau \sim U(\ospace)}|h_{\pa,\tau} - h_{\protp,\tau}|}_{\UF_\gamma(f_\lambda)} = \Expect_{\tau \sim U(\ospace)}\left|\underbrace{\frac{d}{d^2\lambda}[h_{\pa,\tau} - h_{\protp,\tau}]}_{\geq 0 ~ \text{by (\ref{eq:convexity_inside_derivative})}}\right|.
\]
This indicates that $\UF_\gamma(f_\lambda)$ is convex (i.e., we have shown that the second derivative is non-negative).  
\end{proof}

\subsection{Proof of Proposition \ref{prop:repaired_barycenters}}
\paragraph{Proposition \ref{prop:repaired_barycenters}. }\barycenterprop
\begin{proof}
First we recall the definition of geometric repair 
\begin{align*}
 f_{\lambda}(x,\prot) \triangleq (1-\lambda)f(x,\prot) + \lambda \fsdp(x,\prot).
\end{align*}
It is easy to show that for $T_\prot^*$ we have that (wlog) $\fsdp(x,\pa) = T_\prot^*(f(x,\pa))$
\begin{align}
\fsdp(x,\pa) = (\rho_\pa \id + \rho_b T_{\pa}^{\protp}) \circ F_{\mu_\pa}(f(x,\pa)) &= \\ 
 F_{\mustar}^{-1}(F_{\mu_\pa}(f(x,\pa))) &= \\ 
    T_\prot^*(f(x,\pa)). 
\end{align}
Where the first equality follows from Theorem 2.3. in \cite{chzhen2020fairwasserstein}, and the second equality is the definition of $\mustar$. Using this equality in the 
 definition of geometric repair we get
\begin{align}
 f_{\lambda}(x,\prot) &= (1-\lambda)f(x,\prot) + \lambda T_\prot^*(f(x,\prot)) \\ &=
 \left((1-\lambda)\id + \lambda T_\prot^*\right) \circ f(x,\prot) 
\end{align}
If we let $\mu_\prot$ be the groupwise score distribution for group $\prot$ then we know $\mu_\prot = Law(f(\Xrv, \Protrv)|\Protrv = \prot)$ by definition. If we pushfoward $\mu_\prot$ using $\left((1-\lambda)\id + \lambda T_\prot^*\right)$, i.e., 
\begin{align*}
    \left((1-\lambda)\id + \lambda T_\prot^*\right){\# \mu_\prot} = \argmin_{\nu \in \Ppm{2}(\ospace)}  (1-\lambda)\Wass_2^2(\mu_\prot, \nu) + \lambda\Wass_2^2(\mustar, \nu)
\end{align*}
by Claim \ref{claim:mcann} and the uniqueness of $\Wass_2$ barycenters.  Noticing the $\mu_{\prot,\lambda}$ is the score distribution for $f_\lambda(\Xrv,\Protrv)|\Protrv=\prot$ completes the proof. 
\end{proof}

\subsection{Proof of Theorem \ref{prop:pareto_prop}}
\paragraph{Theorem \ref{prop:pareto_prop}. }
\paretooptimal
\begin{proof}
It is clear from the definition of $f_\lambda$ that $\{f_\lambda\}_{\lambda \in \binaryrange}$ forms a pareto front.  Indeed, recall that for any level of unfairness, say $\lambda \UF_\PR(\fsdp)$, that $f_\lambda$ is the regressor which minimizes risk, i.e., 
\begin{align*}
    f_\lambda \leftarrow \arg\min_{\fhat \in \mc F} \mc \lambda R(\fhat) \quad \suchthat \UF_{\PR}(f_\lambda) = (1-\lambda) \UF_{\PR}(\f) . 
\end{align*}
Due to the above, it is easy to see that no classifier can have risk less than $\f_\lambda$, without decreasing $\lambda$, which in turn increase $\UF(\cdot)$, proving the pareto optimality of $\f_\lambda$. Now, suppose for contradiction,  $\{f_\lambda\}_{\lambda \in \binaryrange}$ did not form a pareto front, i.e., there exists some $h \not \in \{f_\lambda\}_{\lambda \in \binaryrange}$ such that $h \succ f_\lambda$  for some $\lambda \in \binaryrange$.  Since $h \succ f_\lambda$ then clearly (WLOG) $\mc R(h) < \mc R(f_\lambda)$.  
However if we select $\lambda_h = \frac{\mc R(h)}{\mc R(f_\lambda)}$ then $\mc R(f_{\lambda_h}) = \mc R(h)$ and subsequently $\UF_\PR(f_{\lambda_h}) = \UF_\PR(h)$, which by definition means $h \in \{f_\lambda\}_{\lambda \in \binaryrange}$.  In the other case where $\UF(h) < \UF(f_\lambda)$ the proof follows identically. In both cases, we arrive at a contradiction indicating that $\{f_\lambda\}_{\lambda \in \binaryrange}$ is indeed a Pareto Frontier. 
\end{proof}

\subsection{Proof of Theorem \ref{thm:empirical_dist}}
\paragraph*{Theorem \ref{thm:empirical_dist}.}\empiricaldisttheorem 
\begin{proof}
To complete this proof, it will be convenient to consider the following mixture distribution
\begin{align*}
    \mc P = \sum_{\prot \in \Prot} \rho_\prot \delta_{\mu_\prot}
\end{align*}
and its empirical variant $\hat{\mc P}$ using $\hat{\rho_\prot}$ and $\hat{\mu}_\prot$.  Relying on the barycenters uniqueness under Assumption \ref{assumption:wass} in $\Wass_2$ (proven by \cite{agueh2011barycenters}) and the consistency of the Wasserstein barycenter \cite{le2017existence}[Theorem 3], proving that $\hat{\mc P} \rightarrow \mc P$ in the Wasserstein Distance is sufficient to prove the convergence $\hat{\mu}_{\prot,\lambda}$.

We now begin the proof. Recall that we can express $\mu_{\prot,\lambda}$ as $\lambda$-weighted barycenter between $\mu_\prot, \mustar$ or as a $\lambda \rho_\protp$ weighted barycenter between $\mu_\pa$ and $\mu_{\protp}$.  Consider the latter formulation, i.e. 
\begin{align*}
    \mc P_\lambda = (1-\lambda)\rho_\protp \delta_{\mu_\pa} + \lambda\rho_\protp \delta_{\mu_\protp}
\end{align*}
Thus via the consistency of Wasserstein barycenters, as stated above, we must only show that $\hat \rho_\prot$ converges to $\rho_\prot$, and that $\hat{\mu}_\prot \rightarrow \mu_\prot$ in $\Wass_2$.  The convergence of $\hat \rho_\prot$ follows by the law of large numbers.  The convergence of $\hat{\mu}_\prot$ follows from the  well known facts that the Wasserstein Distance metrizes the weak convergence of probability measures \cite[Theorem 6.9]{Villani2008OptimalTO}, and that an empirical measure $\hat{\mu}_k \rightarrow \mu$ almost surely, \cite{sampleDistributionsVaradarajan58}.  From these facts it follows that $\Wass_2(\hat{\mu}_\prot, \mu_\prot) \rightarrow 0$ almost surely, completing the proof.    
\end{proof}